\documentclass[10pt,twocolumn,letterpaper]{article}

\usepackage{wacv}
\usepackage{times}
\usepackage{epsfig}
\usepackage{graphicx}
\usepackage{amsmath}
\usepackage{amssymb}

\usepackage{subfigure}
\usepackage{enumitem, rotating}
\usepackage{xspace}
\usepackage{algorithm,algpseudocode}

\usepackage[pagebackref=true,breaklinks=true,letterpaper=true,colorlinks,bookmarks=false]{hyperref}

\wacvfinalcopy %

\def\wacvPaperID{112} %

\newcommand{\pname}{DMC\xspace}

\newcommand{\vct}[1]{\boldsymbol{#1}} %
\newcommand{\field}[1]{\mathbb{#1}}

\ifwacvfinal\fi
\begin{document}

\title{Class-incremental Learning via Deep Model Consolidation}

\author{Junting Zhang$^1$ \quad Jie Zhang$^2$ \quad Shalini Ghosh$^3$ \quad Dawei Li$^3$ \quad Serafettin Tasci$^3$ \quad Larry Heck$^3$\\
	Heming Zhang$^1$ \quad C.-C. Jay Kuo$^1$ \\
	$^1$University of Southern California \quad $^2$Arizona State University \quad $^3$Samsung Research America
\\ \href{mailto:juntingz@usc.edu}{\tt{\small{juntingz@usc.edu}}}
}

\maketitle
\ifwacvfinal\thispagestyle{empty}\fi

\begin{abstract}
Deep neural networks (DNNs) often suffer from ``catastrophic forgetting" during incremental learning (IL) --- an abrupt degradation of performance on the original set of classes when the training objective is adapted to a newly added set of classes. Existing IL approaches tend to produce a model that is biased towards either the old classes or new classes, unless with the help of exemplars of the old data. To address this issue, we propose a class-incremental learning paradigm called Deep Model Consolidation (DMC), which works well even when the original training data is not available. The idea is to first train a separate model only for the new classes, and then combine the two individual models trained on data of two distinct set of classes (old classes and new classes) via a novel double distillation training objective. The two existing models are consolidated by exploiting publicly available unlabeled auxiliary data. This overcomes the potential difficulties due to unavailability of original training data. Compared to the state-of-the-art techniques, DMC demonstrates significantly better performance in image classification (CIFAR-100 and CUB-200) and object detection (PASCAL VOC 2007) in the single-headed IL setting.
\end{abstract}

\section{Introduction}
\label{sec:introduction}

Despite the recent success of deep learning in computer vision for a broad range of tasks \cite{chen2016deeplab, huang2017densely, krizhevsky2012imagenet, lin2018focal}, classical training paradigm of deep models is ill-equipped for incremental learning (IL). Most deep neural networks can only be trained when the complete dataset is given and all classes are known prior to training. However, the real world is dynamic and new categories of interest can emerge over time. Re-training a model from scratch whenever a new class is encountered can be prohibitively expensive due to training data storage requirements and the computational cost of full retrain. 
Directly fine-tuning the existing model on only the data of new classes using stochastic gradient descent (SGD) optimization is not a better approach either, as this might lead to the notorious \textit{catastrophic forgetting} problem~\cite{goodfellow2013empirical,mccloskey1989catastrophic}, which can result in severe performance degradation on old tasks. 

\begin{figure}[tbp]
\begin{center}
   \includegraphics[width=\linewidth]{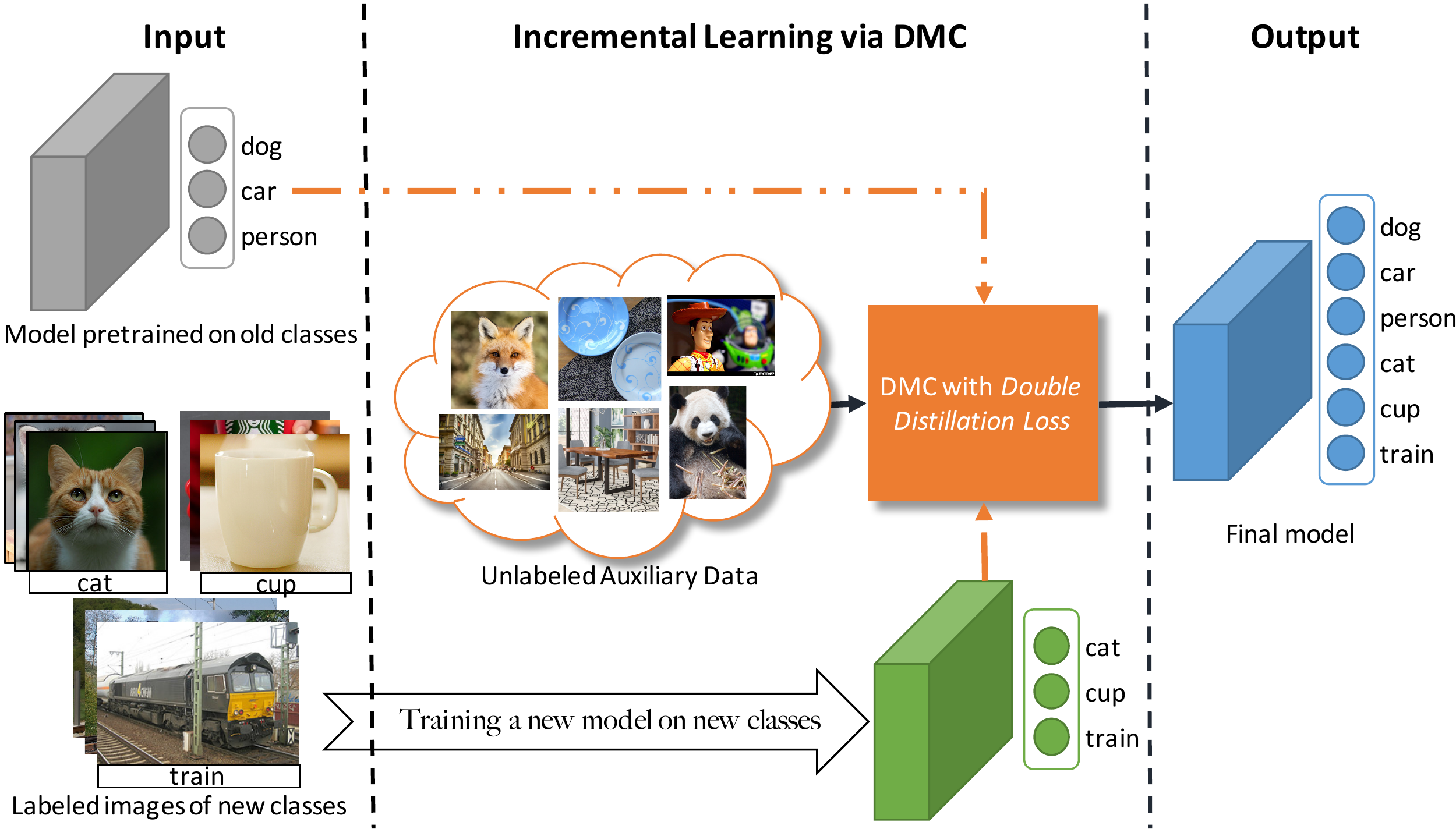}
\end{center}
	\caption{Overview of the proposed incremental learning algorithm. Given a model pretrained on existing classes and labeled data of new classes, we first train a new model for recognizing instances of new classes; we then combine the old model and the new model using the novel deep model consolidation (\pname) module, which leverages external unlabeled auxiliary data. The final model suffers less from forgetting the old classes, and achieves high recognition accuracy for the new classes. }
	\label{fig:overview}
\end{figure}

We consider a realistic, albeit strict and challenging, setting of class-incremental learning, where the system must satisfy the following constraints: 1) the original training data for old classes are no longer accessible when learning new classes --- this could be due to a variety of reasons, \eg, legacy data may be unrecorded, proprietary, too large to store, or subject to privacy constraint when training the model for a new task; this is a practical concern in various academic and industrial applications, where the model can be transferred from one party to another but data should be kept private, and a practitioner wants to augment the model to learn new classes; 2) the system should provide a competitive multi-class classifier for the classes observed so far, \ie single-headed classification should be supported, which does not require any prior information of the test data; 3) the model size should remain relatively unchanged after learning new classes.

Several attempts have been made to enable IL for DNNs, but none of them satisfies all of these constraints. Some recent works \cite{castro2018end,chaudhry2019efficient,he2018exemplar,javed2018revisiting,lopez2017gradient,rebuffi2017icarl} that rely on the storage of partial old data have made impressive progress. They are arguably not memory efficient and storing data for the life time involves violate some practical constraints such as copyright or privacy issues, which is common in the domains like bio-informatics \cite{samet2013incremental}.
The performance of the existing methods that do not store any past data is yet unsatisfactory. Some of these methods rely on 
incrementally training generative models \cite{kemker2018fearnet, shin2017continual}, which is a harder problem to solve; while others fine-tune the old model on the new data with certain regularization techniques to prevent forgetting \cite{aljundi2018memory,chaudhry2018riemannian,kirkpatrick2017overcoming,li2017learning,shmelkov2017incremental,zenke2017continual,zhang2020regularize}. We argue that the ineffectiveness of these regularization-based methods is mainly due to the \emph{asymmetric information} between old classes and new classes in the fine-tuning step. New classes have explicit and strong supervisory signal from the available labeled data, whereas the information for old classes is implicitly given in the form of a noisy regularization term. Moreover, if we over-regularize the model, the model will fail to adapt to the new task, which is referred to as \emph{intransigence} \cite{chaudhry2018riemannian} in the IL context. As a result, these methods have intrinsic bias towards either the old or the new classes in the final model, and it is extremely difficult to find a sweet spot considering that in practice we do not have a validation dataset for the old classes during incremental learning.

 As depicted in Fig.~\ref{fig:overview}, we propose a novel paradigm for class-incremental learning called deep model consolidation (\pname), which first trains a separate model for the new classes using labeled data, and then combines the new and old models using publicly available unlabeled auxiliary data via a novel double distillation training objective. \pname eliminates the intrinsic bias caused by the information asymmetry or over-regularization in the training, as the proposed double distillation objective allows the final student model to learn from two teacher models (the old and new models) simultaneously. \pname overcomes the difficulty introduced by loss of access to legacy data by leveraging unlabeled auxiliary data, where the abundant transferable representations are mined to facilitate IL. Furthermore, using the auxiliary data rather than the training data of the new classes ensures the student model absorbs the knowledge transferred from the both teacher models in an unbiased way.

Crucially, we do not require the auxiliary data share the class labels or generative distribution of the target data. The only requirement is that they are generic, diversified, and generally related to the target data. 
Usage of such unlabeled data incurs no additional dataset construction and maintenance cost since they can be crawled from the web effortlessly when needed and discarded once the IL of new classes is complete. Furthermore, note that the symmetric role of the two teacher models in \pname has a valuable extra benefit in the generalization of our method; it can be directly applied to combine any two arbitrary pre-trained models that can be downloaded from the Internet for easy deployment (\ie, only one model needs to be deployed instead of two), without access to the original training data.

To summarize, our main contributions include:
\setlist{nolistsep}
\begin{itemize}[noitemsep]
    \item A novel paradigm for incremental learning which exploits external unlabeled data, which can be obtained at negligible cost. This is an illuminating perspective for IL, which bypasses the constraint of having old data stored by finding some cheap substitute that does not need to be stored.
    \item A new training objective function to combine two deep models into one single compact model to promote symmetric knowledge transfer. The two models can have different architectures, and they can be trained on data of distinct set of classes.
    \item An approach to extend the proposed paradigm to incrementally train modern one-stage object detectors, to which the existing methods are not applicable.
    \item Extensive experiments that demonstrate the substantial performance improvement of our method over existing approaches on large-scale image classification and object detection benchmarks in the IL setting.
\end{itemize}

\section{Related work}
\label{sec:related_work}
McCloskey \etal \cite{mccloskey1989catastrophic} first identified the \textit{catastrophic forgetting} effect in the connectionist models, where the memory about the old data is overwritten when retraining a neural network with new data. Recently, researchers have been actively developing methods to alleviate this effect.

\noindent\textbf{Regularization methods.} Regularization methods enforce additional constraints on the weight update, so that the new concepts are learned while retaining the prior memories. Goodfellow \etal \cite{goodfellow2013empirical} found that dropout \cite{srivastava2014dropout} could reduce forgetting for multi-layer perceptrons sometimes. One line of work constrains the network parameters that are important to the old tasks to stay close to their old values, while looking for a solution to a new task in the neighborhood of the old one. EWC~\cite{kirkpatrick2017overcoming} and its variants~\cite{chaudhry2018riemannian,schwarz2018progress} use Fisher information matrix to estimate the weight importance; MAS~\cite{aljundi2018memory} uses the gradients of the network output; SI~\cite{zenke2017continual} uses the path integral over the optimization trajectory instead. RWalk~\cite{chaudhry2018riemannian} combines EWC~\cite{kirkpatrick2017overcoming} and SI~\cite{zenke2017continual}. Information about the old task and new task is not symmetric during learning in these methods; besides, the network may become stiffer and stiffer to adapt to the new task as it learns more tasks over time. 
Li and Hoiem \cite{li2017learning} pursued another direction by proposing the Learning without Forgetting (LwF) method, which finetunes the network using the images of new classes with knowledge distillation \cite{hinton2015distilling} loss, to encourage the output probabilities of old classes for each image to be close to the original network outputs. However, information asymmetry between old classes and new classes still exists. Image samples from new data may severely deviate from the true distribution of the old data, which further aggravates the information asymmetry. Instead, we assign two teacher models to one student network to guarantee the symmetric information flow from old- and new-class models into the final model. IMM~\cite{lee2017overcoming} first finetunes the network on the new task with regularization, and then blends the obtained model with the original model through moment matching. Though conceptually similar, our work is different from IMM~\cite{lee2017overcoming} in the following ways: 1) we do not use regularized-finetuning from old-class model when training the new model for the new classes, so we can avoid intrinsic bias towards the old classes and suboptimal solution for the new task; 2) we do not assume the final posterior distribution for all the tasks is Gaussian, which is a strong assumption for DNNs.

\noindent\textbf{Dynamic network methods.} Dynamic network methods~\cite{mallya2018piggyback,mallya2018packnet,serra2018overcoming,yoon2018lifelong} dedicate a part of the network or a unique feed-forward pathway through neurons for each task. At test time, they require the task label to be specified to switch to the correct state of the network, which is not applicable in the class-IL where task labels are not available.

\noindent\textbf{Rehearsal and pseudo-rehearsal methods.} In rehearsal methods \cite{castro2018end,chaudhry2019efficient,javed2018revisiting,lopez2017gradient,nguyen2017variational,rebuffi2017icarl}, past information is periodically replayed to the model to strengthen memories it has already learned, which is done by interleaving data from earlier sessions with the current session data \cite{robins1995catastrophic}. However, storage of past data is not resource efficient and may violate some practical constraints such as copyright or privacy issues. Pseudo-rehearsal methods attempt to alleviate this issue by using generative models to generate pseudopatterns \cite{robins1995catastrophic} that are combined with the current samples. However, this requires training a generative model in the class-incremental fashion, which is an even harder problem to solve. Existing such methods do not produce competitive results \cite{kemker2018fearnet,shin2017continual} unless supported by real exemplars \cite{he2018exemplar}.

\noindent\textbf{Incremental learning of object detectors.} Shmelkov \etal \cite{shmelkov2017incremental} adapted LwF for the object detection task. However, their framework can only be applied to object detectors in which proposals are computed externally, \eg, Fast R-CNN \cite{girshick2015fast}. In our experiments, we show that our method is applicable to more efficient modern single-shot object detection architectures, \eg, RetinaNet \cite{lin2018focal}.

\noindent\textbf{Exploiting external data.} In computer vision, the idea of employing external data to improve performance of a target task has been explored in many contexts. Inductive transfer learning \cite{csurka2017comprehensive,zhang2018fully} aims to transfer and reuse knowledge in labeled out-of-domain instances. Semi-supervised learning \cite{chapelle2009semi,zhu2006semi} attempts to exploit the usefulness of unlabeled in-domain instances. Our work shares a similar spirit with self-taught learning \cite{raina2007self}, where we use unlabeled auxiliary data but do not require the auxiliary data to have the same class labels or generative distribution as the target data. Such unlabeled data is significantly easier to obtain compared to typical semi-supervised or transfer learning settings.

\section{Method}
\label{sec:method}
Let's first formally define the class-incremental learning setting. Given a labeled data stream of sample sets $X^1, X^2, \cdots$, where $X^y = \{x^y_1, \cdots x^y_{y_n} \}$ denotes the samples of class $y \in \field{N}^{+}$, we learn one class or group of classes at a time. During each learning session, we only have training data $\mathcal{D}_{new} = \{X^{s+1},\dots, X^{t}\}$ of newly available classes $s+1,\cdots,t$, while the training data of the previously learned classes $\{X^{1},\dots, X^{s}\}$ are no longer accessible. However, we have the model obtained in the previous session, which is an $s$-class classifier $f_{old}(x; \Theta_{old})$. Our goal is to train a $t$-class classifier $f(x; \Theta)$ without catastrophic forgetting on old classes or significant underperformance on the new classes. We assume that all models are implemented as DNNs where $x$ and $\Theta$ denote the input and the parameters of the network, respectively. 

We perform IL in two steps: the first step is to train a $(t-s)$-class classifier using training data $\mathcal{D}_{new}$, which we refer as the \emph{new} model $f_{new}(x; \Theta_{new})$; the second step is to consolidate the old model and the new model.

The new class learning step is a regular supervised learning problem and it can be solved by standard back-propagation. The model consolidation step is the major contribution of our work, where we propose a method called Deep Model Consolidation (\pname) for image classification which we further extend to another classical computer vision task, object detection.

\subsection{\pname for image classification}
\label{ssec:classification}

We start by training a new CNN model $f_{new}$ on new classes using the available training data $D_{new}$ with standard softmax cross-entropy loss. Once the new model is trained, we have two CNN models specialized in classifying either the old classes or the new classes. After that, the goal of the consolidation is to have a single compact model that can perform the tasks of both the old model and the new model simultaneously. Ideally, we have the following objective:
\begin{equation}
f(x; \Theta)[j] = \begin{cases}
f_{old}(x; \Theta_{old})[j] , & 1 \leq j \leq s\\
f_{new}(x; \Theta_{new})[j], & s < j \leq t
\end{cases},\forall x \in \mathcal{I}
\end{equation}
where $j$ denotes the index of the classification score associated with $j$-th class, and $\mathcal{I}$ denotes the joint distribution from which samples of class $1, \cdots, t$ are drawn. We want the output of the consolidated model to approximate the combination of the network outputs of the old model and the new model. To achieve this, the network response of the old model and the new model is employed as supervisory signals in joint training of the consolidated model.

Knowledge distillation (KD) \cite{hinton2015distilling} is a popular technique to transfer knowledge from one network to another. Originally, KD was proposed to transfer knowledge from a cumbersome network to a light-weight network performing the same task, and no novel class was introduced. We generalize the basic idea in KD and propose a \textit{double distillation loss} to enable class-incremental learning. Here, we define the \emph{logits} as the inputs to the final softmax layer. We run a feed-forward pass of $f_{old}$ and $f_{new}$ for each training image, and collect the logits of the two models $\vct{\hat{y}}_{old}= [\hat{y}^1,\cdots,\hat{y}^s]$ and $\vct{\hat{y}}_{new}=[\hat{y}^{s+1},\cdots,\hat{y}^t]$, respectively, where the superscript is the class label associated with the neuron in the model. Then we minimize the difference between the logits produced by the consolidated model and the combination of logits generated by the two existing specialist models, according to some distance metric. We choose $L_2$ loss \cite{ba2014deep} as the distance metric because it demonstrates stable and good performance, see  \S~\ref{sssec:loss} for discussions.

Due the absence of the legacy data, we cannot consolidate the two models using the old data. Thus some auxiliary data has to be used. 
If we assume that natural images lie on an ideal low-dimensional manifold,
we can approximate the distribution of our target data via sampling from readily available unlabeled data from a similar domain. Note that the auxiliary data do not have to be stored persistently; they can be crawled and fed in mini-batches on-the-fly in this stage, and discarded thereafter.

Specifically, the training objective for consolidation is 
\begin{equation}
\min_{\Theta} \frac{1}{|\mathcal{U}|}\sum_{x_i \in \mathcal{U}} L_{dd}(\vct{y}_i, \vct{\mathring{y}}_i),
\end{equation}
where $\mathcal{U}$ denotes the unlabeled auxiliary training data, and the double distillation loss $L_{dd}$ is defined as:
\begin{equation}
L_{dd}(\vct{y}, \vct{\mathring{y}}) = \frac{1}{t}\sum_{j=1}^{t} \left(y^j - \mathring{y}^j\right)^2,
\label{eq:cls_loss}
\end{equation}
in which $y^j$ is the logit produced by the consolidated model for the $j$-th class, and
\begin{equation}
    \mathring{y}^j =  \begin{cases}
    \hat{y}^j - \frac{1}{s}\sum_{k=1}^s\hat{y}^k, & 1\leq j \leq s \\
    \hat{y}^j - \frac{1}{t-s}\sum_{k=s+1}^t\hat{y}^k, & s < j \leq t \\
    \end{cases}
    \label{eq:normalize}
\end{equation}
where $\vct{\hat{y}}$ is the concatenation of $\vct{\hat{y}}_{old}$ and $\vct{\hat{y}}_{new}$.

The regression target $\vct{\mathring{y}}$ is the concatenation of normalized logits of the two specialist models. We normalize $\vct{\hat{y}}$ by subtracting its mean over the class dimension (Eq. \ref{eq:normalize}). This serves as a step of bias calibration for the two set of classes. It unifies the scale of logits produced by the two models, but retains the relative magnitude among the classes, so that the symmetric information flow can be enforced.

Notably, to avoid the intrinsic bias toward either old or new classes, $\Theta$ should not be initialized from $\Theta_{old}$ or $\Theta_{new}$; we should also avoid the usage of training data for the new classes $\mathcal{D}_{new}$ in the model consolidation stage.

\subsection{\pname for object detection}
We extend the IL approach given in Section~\ref{ssec:classification} for modern one-stage object detectors, which are nearly as accurate as two-stage detectors but run much faster than the later ones.
A single-stage object detector divides the input image into a fixed-resolution 2D grid (the resolution of the grid can be multi-level), where higher resolution means that the area corresponding to the image region (\emph{i.e.}, receptive field) of each cell in the grid is smaller. There are a set of bounding-box templates with fixed sizes and aspect ratios, called anchor boxes, which are associated with each spatial cell in the grid. Anchor boxes serve as references for the subsequent prediction. The class label and the bounding box location offset relative to the anchor boxes are predicted by the classification subnet and bounding boxes regression subnet, respectively, which are shared across all the feature pyramid levels \cite{lin2016feature}. 

In order to apply \pname to incrementally train an object detector, we have to consolidate the classification subnet and bounding boxes regression subnet simultaneously. Similar to the image classification task, we instantiate a new detector whenever we have training data $\mathcal{D}_{new}$ for new object classes. After the new detector is properly trained, we then use the outputs of the two specialist models to supervise the training of the final model.

\noindent\textbf{Anchor boxes selection.} In one-stage object detectors, a huge number of anchor boxes have to be used to achieve decent performance. For example, in RetinaNet~\cite{lin2018focal}, $\sim$100k anchor boxes are used for an image of resolution $800 \times 600$. Selecting a smaller number of anchor boxes speeds up forward-backward pass in training significantly. The naive approach of randomly sampling some anchor boxes doesn't consider the fact that the ratio of positive anchor boxes and negative ones is highly imbalanced, and negative boxes that correspond to background carry little information for knowledge distillation. In order to efficiently and effectively distill the knowledge of the two teacher detectors in the \pname stage, we propose a novel anchor boxes selection method to selectively enforce the constraint for a small set of anchor boxes. For each image sampled from the auxiliary data, we first rank the anchor boxes by the objectness scores. The objectness score ($os$) for an anchor box is defined as: \begin{equation}
os \triangleq \max\{p^1, \cdots, p^s, p^{s+1},\cdots,p^t\},
\end{equation} where $p^1, \cdots, p^s$ are classification probabilities produced by the old-class model, and $p^{s+1},\cdots,p^t$ are from the new-class model. Intuitively, a high objectness score for a box implies a higher probability of containing a foreground object. The predicted classification probabilities of the old classes are produced by the old model, and new classes by the new model. We use the subset of anchor boxes that have the highest objectness scores and ignore the others.

\noindent\textbf{\pname for classification subnet.} Similar to the image classification case in Sec. \ref{ssec:classification}, for each selected anchor box, we calculate the double distillation loss between the logits produced by the classification subnet of the consolidated model $\vct{y}$ and the normalized logits generated by the two existing specialist models $\vct{\mathring{y}}$. The loss term of \pname for the classification subnet $L_{cls}(\vct{y}, \vct{\mathring{y}})$ is identical to Eq. \ref{eq:cls_loss}. 

\noindent\textbf{\pname for bounding box regression subnet.} The output of the bounding box regression subnet is a tuple of spatial offsets $\vct{t} = (t_x, t_y, t_h, t_w)$, which specifies a scale-invariant translation and log-space height/width shift relative to an anchor box. For each anchor box selected, we need to set its regression target properly. If the class that has the highest predicted class probability is one of the old classes, we choose the old model's output as the regression target, otherwise, the new model's output is chosen. In this way, we encourage the predicted bounding box of the consolidated model to be closer to the predicted bounding box of the most probable object category. Smooth $L_1$ loss \cite{girshick2015fast} is used to measure the closeness of the parameterized bounding box locations. The loss term of \pname for the bounding box regression subnet is as follows:
\begin{equation}
L_{loc}(\vct{t}, \vct{\hat{t}}) = \sum_{k={x,y,h,w}} smooth_{L_1}(t_k - \hat{t}_k),
\end{equation}
in which
\begin{equation}
    \vct{\hat{t}} = \begin{cases}
\vct{\hat{t}}_{old} , & \max_{1 \leq j \leq s} \hat{y}^j > \max_{s < j \leq t} \hat{y}^j \\
\vct{\hat{t}}_{new}, & \text{otherwise}
\end{cases},
\end{equation}

\noindent\textbf{Overall training objective.} The overall \pname objective function for the object detection is defined as
\begin{equation}
\label{eq:det_loss}
    \min_{\Theta} \frac{1}{|\mathcal{U}|}\sum_{x_i \in \mathcal{U}} L_{cls}(\vct{y}_i, \vct{\mathring{y}}_i) + \lambda L_{loc}( \vct{t}_i, \vct{\hat{t}}_i)
\end{equation}
where $\lambda$ is a hyper-parameter to balance the two loss terms.
\section{Experiments}
\label{sec:experiments}
\subsection{Evaluation protocols}
There are two evaluation protocols for incremental learning. In one setting, the network has different classification layers (multiple “heads”) for each task, where each head can differentiate the classes learned only in this task; it relies on an oracle to decide on the task at test time, which would result in a misleading high test accuracy \cite{chaudhry2018riemannian, liu2018rotate}. In this paper, we adopt a practical yet challenging setting, namely ``single-head" evaluation, where the output space consists of all the $t$ classes learned so far, and the model has to learn to resolve the confusion among the classes from different tasks, when task identities are not available at test time.

\subsection{Incremental learning of image classifiers}
\subsubsection{Experimental setup}
We evaluate our method on iCIFAR-100 benchmark as done in iCaRL \cite{rebuffi2017icarl}, which uses CIFAR-100 \cite{krizhevsky2009learning} data and learn all 100 classes in groups of $g = 5, 10, 20$ or $50$ classes at a time. The evaluation metric is the standard top-1 multi-class classification accuracy on the test set. For each experiment, we run this benchmark 5 times with different class orderings and then report the averages and standard deviations of the results. We use ImageNet32$\times$32 dataset \cite{chrabaszcz2017downsampled} as the source of auxiliary data in the model consolidation stage. The images are down-sampled versions of images from ImageNet ILSVRC \cite {imagenet_cvpr09, ILSVRC15} training set. We exclude the images that belong to the CIFAR-100 classes, which results in 1,082,340 images. Following iCaRL \cite{rebuffi2017icarl}, we use a 32-layer ResNet \cite{he2016deep} for all experiments and the model weights are randomly initialized.

 \subsubsection{Experimental results and discussions}
We compare our method against the state-of-the-art exemplar-free incremental learning methods \textit{EWC++}~\cite{chaudhry2018riemannian,kirkpatrick2017overcoming}, \textit{LwF}~\cite{li2017learning}, \textit{SI}~\cite{zenke2017continual}, \textit{MAS}~\cite{aljundi2018memory}, \textit{RWalk}~\cite{chaudhry2018riemannian} and some baselines with $g=5, 10, 20, 50$. \textit{Finetuning} denotes the case where we directly fine-tune the model trained on the old classes with the labeled images of new classes, without any special treatment for catastrophic forgetting. \textit{Fixed Representation} denotes the approach where we freeze the network weights except for the classification layer (the last fully connected layer) after the first group of classes has been learned, and we freeze the classification weight vector after the corresponding classes have been learned, and only fine-tune the classification weight vectors of new classes using the new data. This approach usually underfits for the new classes due to the limited degree of freedom and incompatible feature representations of the frozen base network. \textit{Oracle} denotes the upper bound results via joint training with all the training data of the classes learned so far.

\begin{figure}[htbp]
	\begin{center}
	\includegraphics[width=0.49\linewidth]{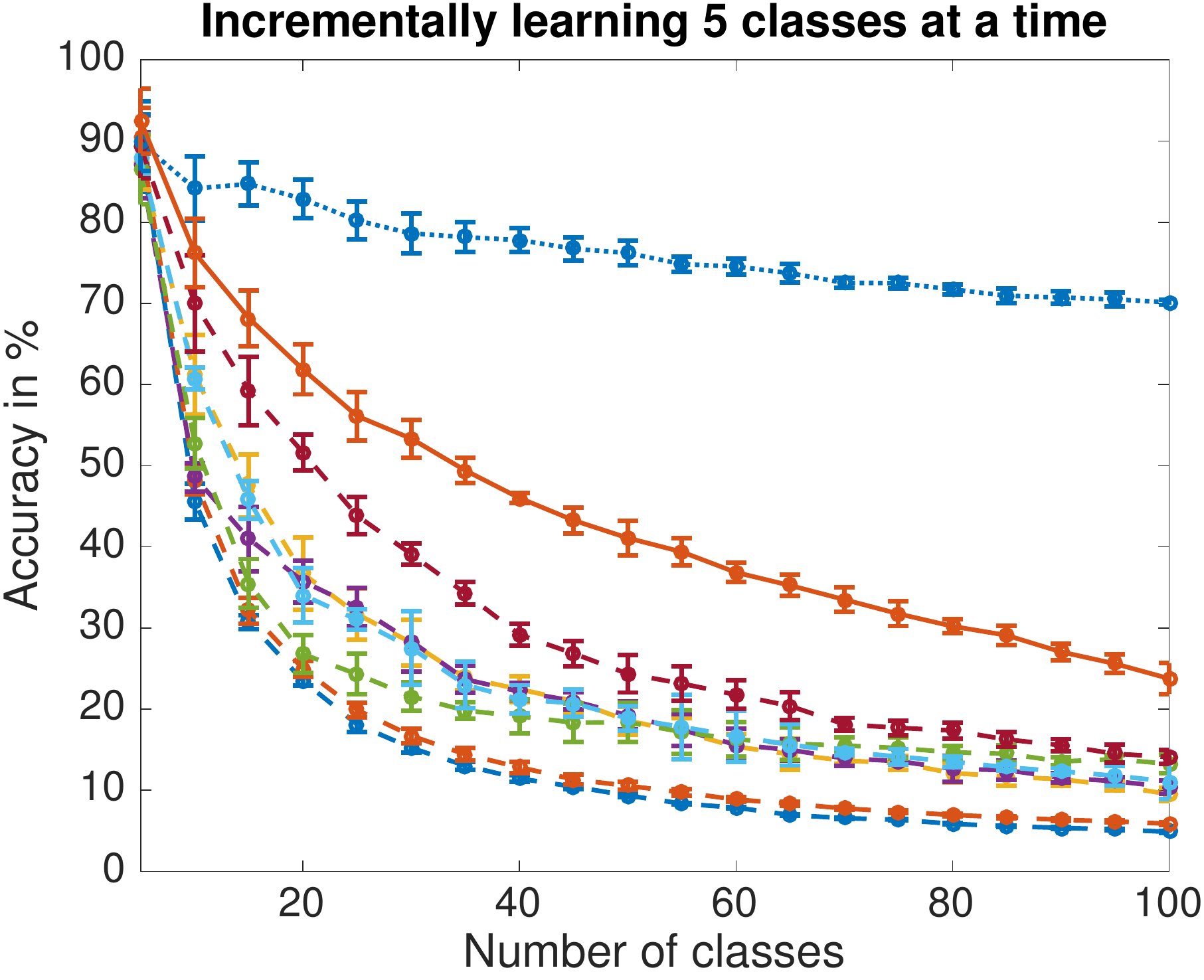}
	\includegraphics[width=0.49\linewidth]{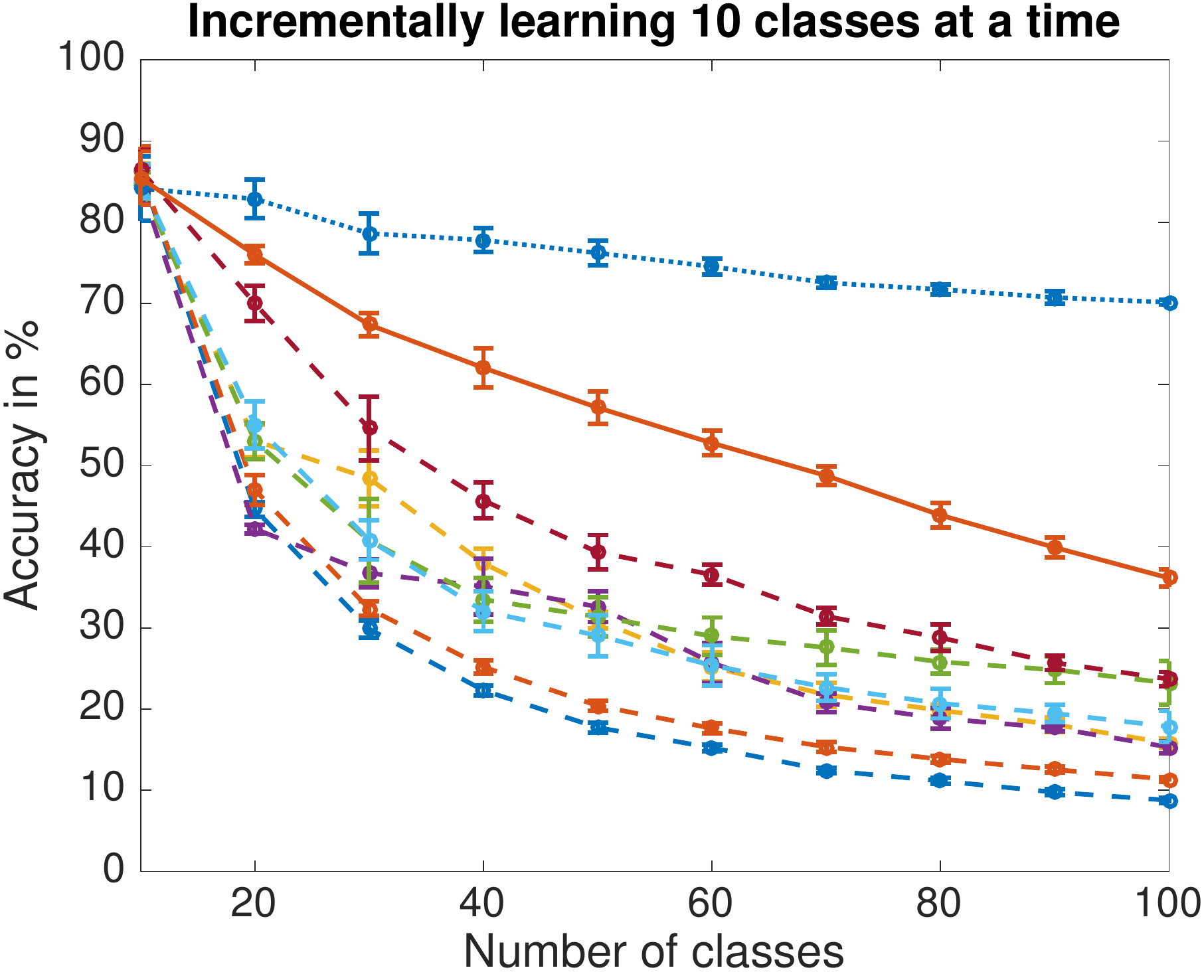}
	\includegraphics[width=0.49\linewidth]{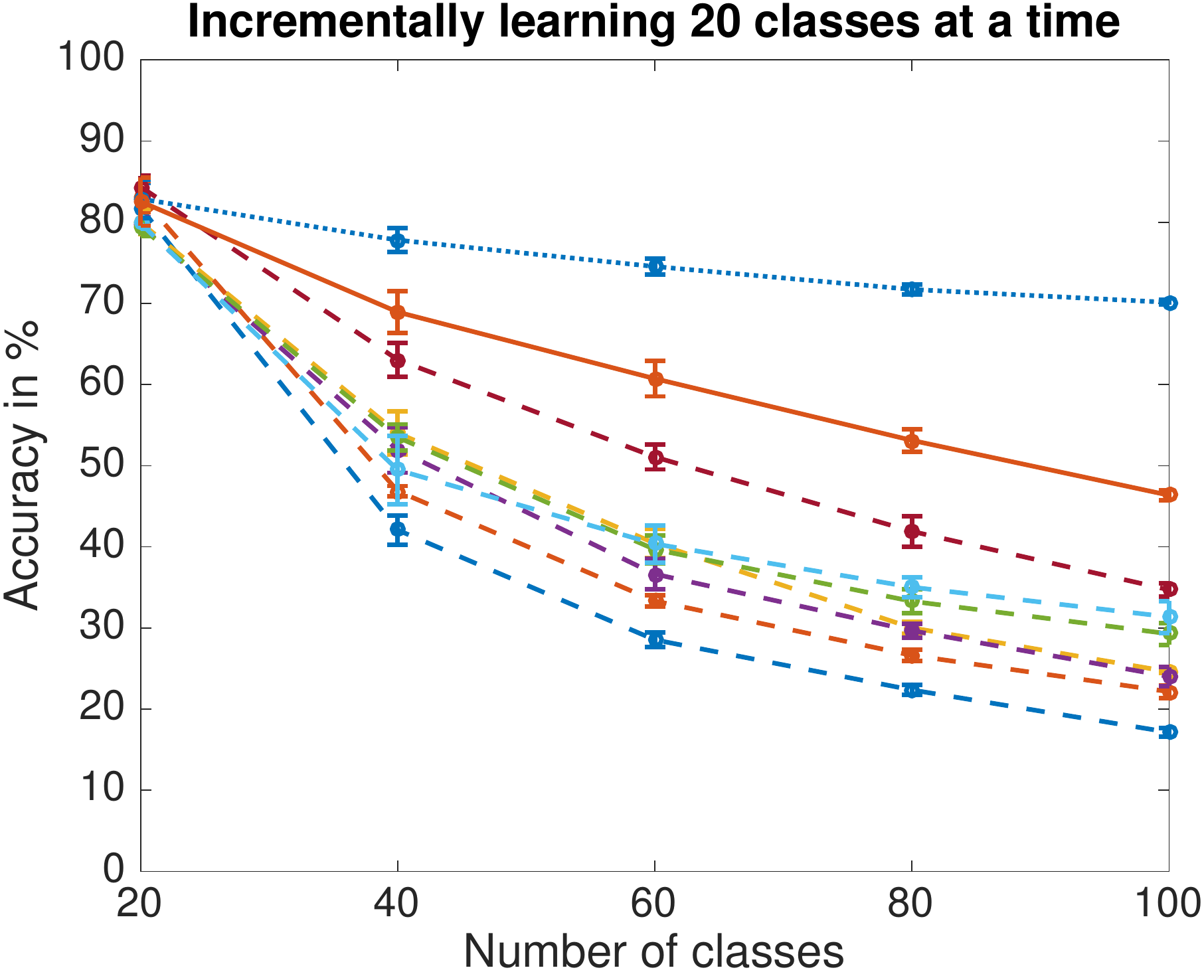}
	\includegraphics[width=0.49\linewidth]{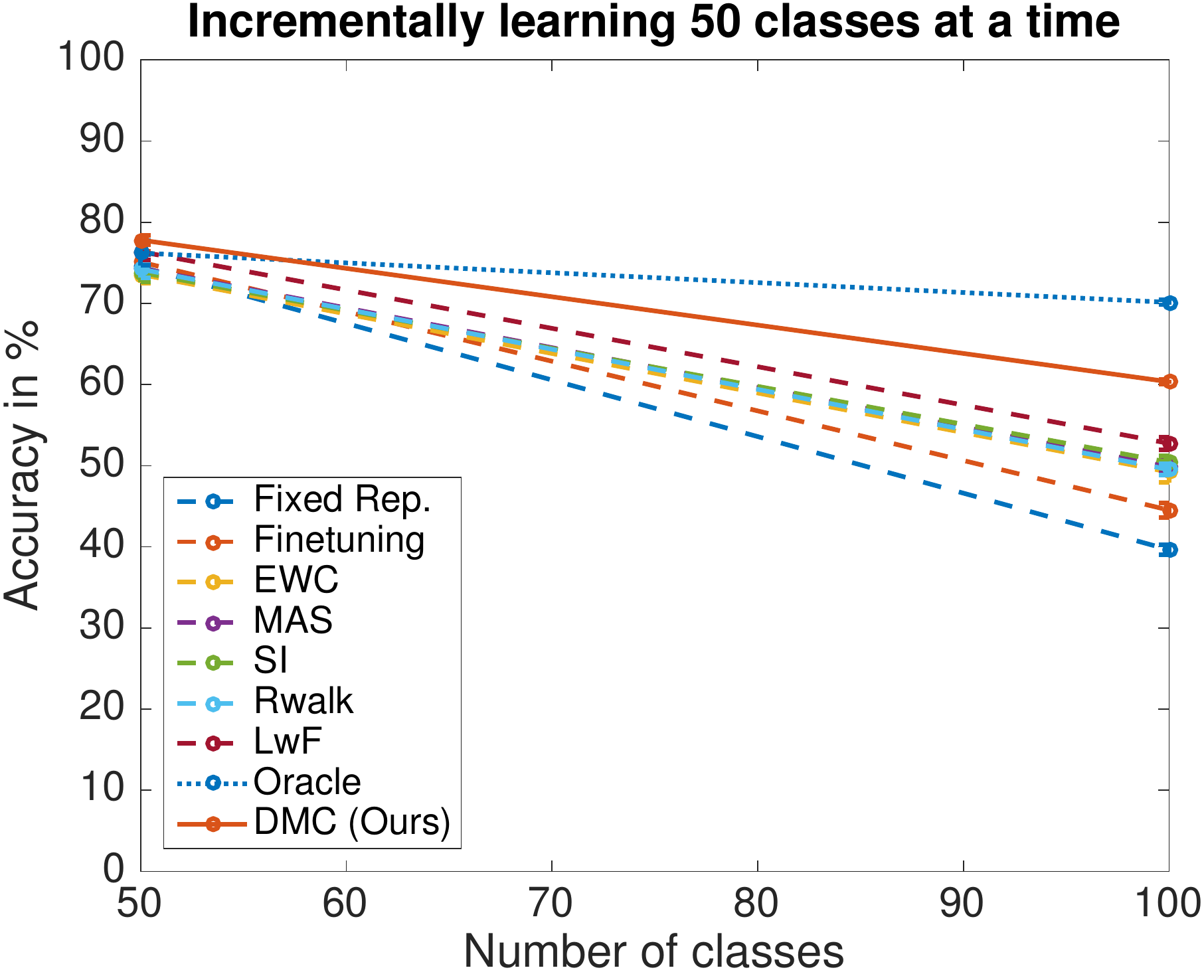}
	\end{center}
	\caption{Incremental learning with group of $g=5,10,20,50$ classes at a time on iCIFAR-100 benchmark.}
	\label{fig:cifar100}
\end{figure}

\begin{figure}[hbt]
	\centering
	\includegraphics[width=\linewidth]{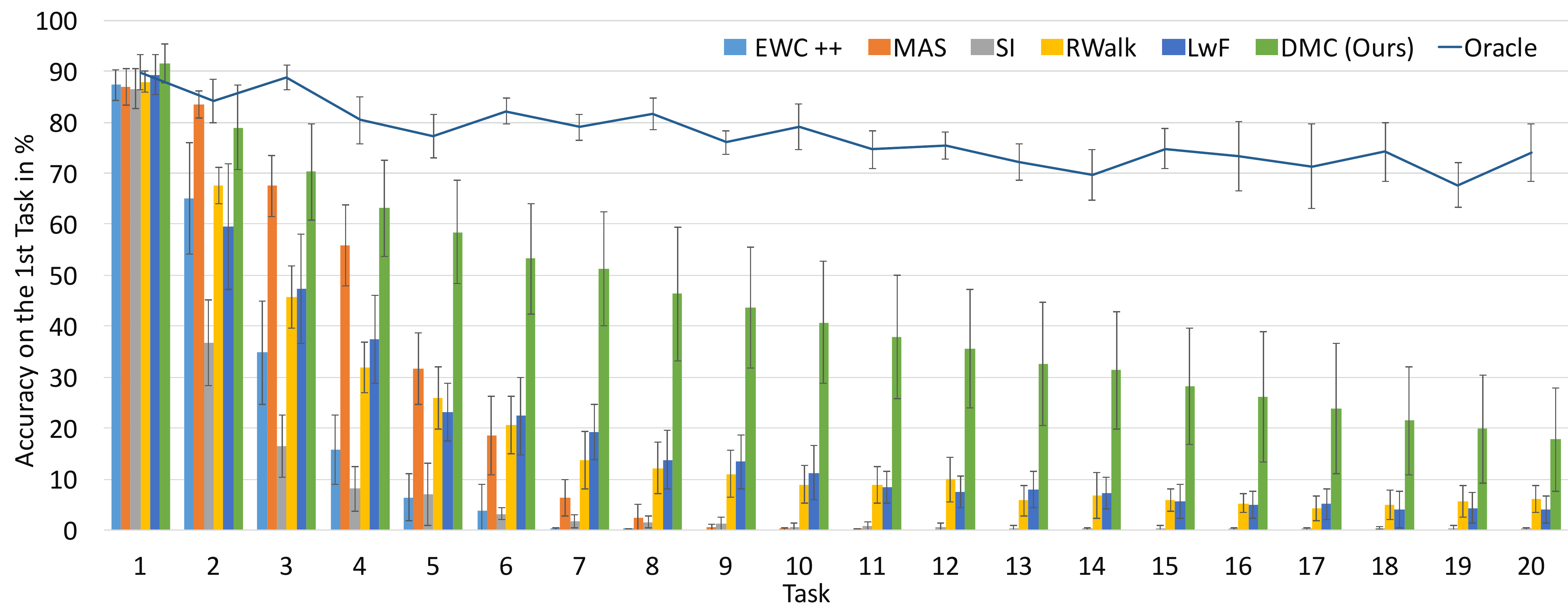}
	\caption{Performance variation on the first task when trained incrementally over 20 tasks ($g=5$) on iCIFAR-100.}
	\label{fig:acc_on_1st_task}
\end{figure}

\begin{figure}[htbp]
	\centering
	\subfigure[\pname]{\includegraphics[width=0.23\linewidth]{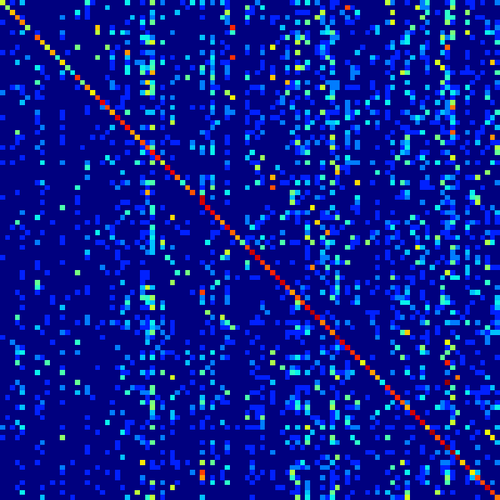}}
	\subfigure[LwF]{
		\label{fig:cm_lwf}
		\includegraphics[width=0.23\linewidth]{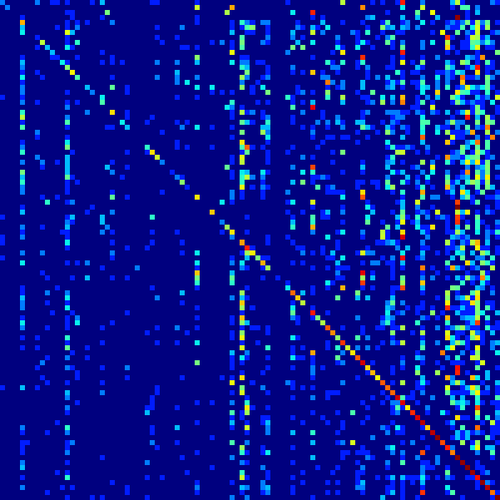}}
	\subfigure[Finetuning]{
		\label{fig:cm_ft}
		\includegraphics[width=0.23\linewidth]{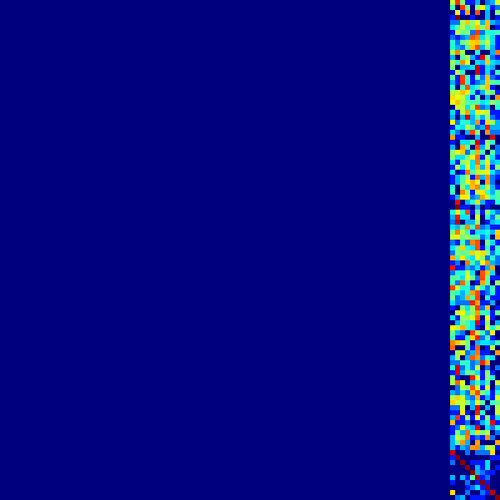}}
	\subfigure[Fixed Repr.]{
		\label{fig:cm_fr}
		\includegraphics[width=0.23\linewidth]{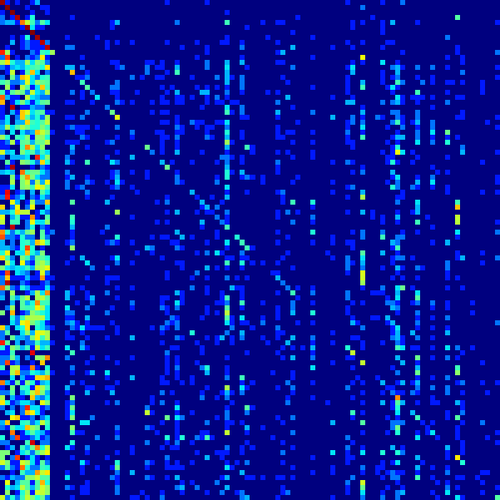}}
	\caption{Confusion matrices of methods on iCIFAR-100 when incrementally learning 10 classes in a group. The entries transformed by $log(1+x)$ for better visibility. Fig. \ref{fig:cm_lwf}, \ref{fig:cm_ft} and \ref{fig:cm_fr} are from \cite{rebuffi2017icarl}. (Best viewed in color.)}
	\label{fig:cifar100-cm}
\end{figure}

The results are shown in Fig.~\ref{fig:cifar100}.  Our method outperforms all the methods by a significant margin across all the settings consistently. We used the official code\footnote{\href{https://github.com/facebookresearch/agem}{https://github.com/facebookresearch/agem}} for \cite{chaudhry2018riemannian} to get the results for \textit{EWC++}~\cite{chaudhry2018riemannian,kirkpatrick2017overcoming}, \textit{SI}~\cite{zenke2017continual},  \textit{MAS}~\cite{aljundi2018memory} and \textit{RWalk}~\cite{chaudhry2018riemannian}. We found they are highly sensitive to the hyperparameter that controls the strength of regularization due to the asymmetric information between old classes and new classes, so we tune the hyperparameter using a held-out validation set for each setting separately, and report the best result for each case. The results of \textit{LwF}~\cite{li2017learning} are from iCaRL~\cite{rebuffi2017icarl} and they are the second-best in all the settings.

It can be also observed that \pname demonstrates a stable performance across different $g$, in contrast to other regularization-based methods, where the disadvantages of inherent asymmetric information flow reveal more, as we incrementally learn more sessions. They struggle in finding the good trade-off between forgetting and intransigence.

Fig. \ref{fig:acc_on_1st_task} illustrates how the accuracy on the first group of classes changes as we learn more and more classes over time. While the previous methods \cite{aljundi2018memory,chaudhry2018riemannian,kirkpatrick2017overcoming,li2017learning,zenke2017continual} all suffer from catastrophic forgetting on the first task, \pname shows considerably more gentle slop of the forgetting curve. Though the standard deviations seems high, which is due to the random class ordering in each run, the relative standard deviations (RSD) are at a reasonable scale for all methods.

We visualize the confusion matrices of some of the methods in Fig. \ref{fig:cifar100-cm}. \textit{Finetuning} forgets old classes and makes predictions based only on the last learned group. \textit{Fixed Representation} is strongly inclined to predict the classes learned in the first group, on which its feature representation is optimized. The previous best performing method \textit{LwF} does a better job, but still has many more non-zero entries on the recently learned classes, which shows strong evidence of information asymmetric between old classes and new classes. On the contrary, the proposed \pname shows a more homogeneous confusion matrix pattern and thus has visibly less intrinsic bias towards or against the classes that it encounters early or late during learning.

\begin{figure}[htb]
	\centering
	\includegraphics[width=0.49\linewidth]{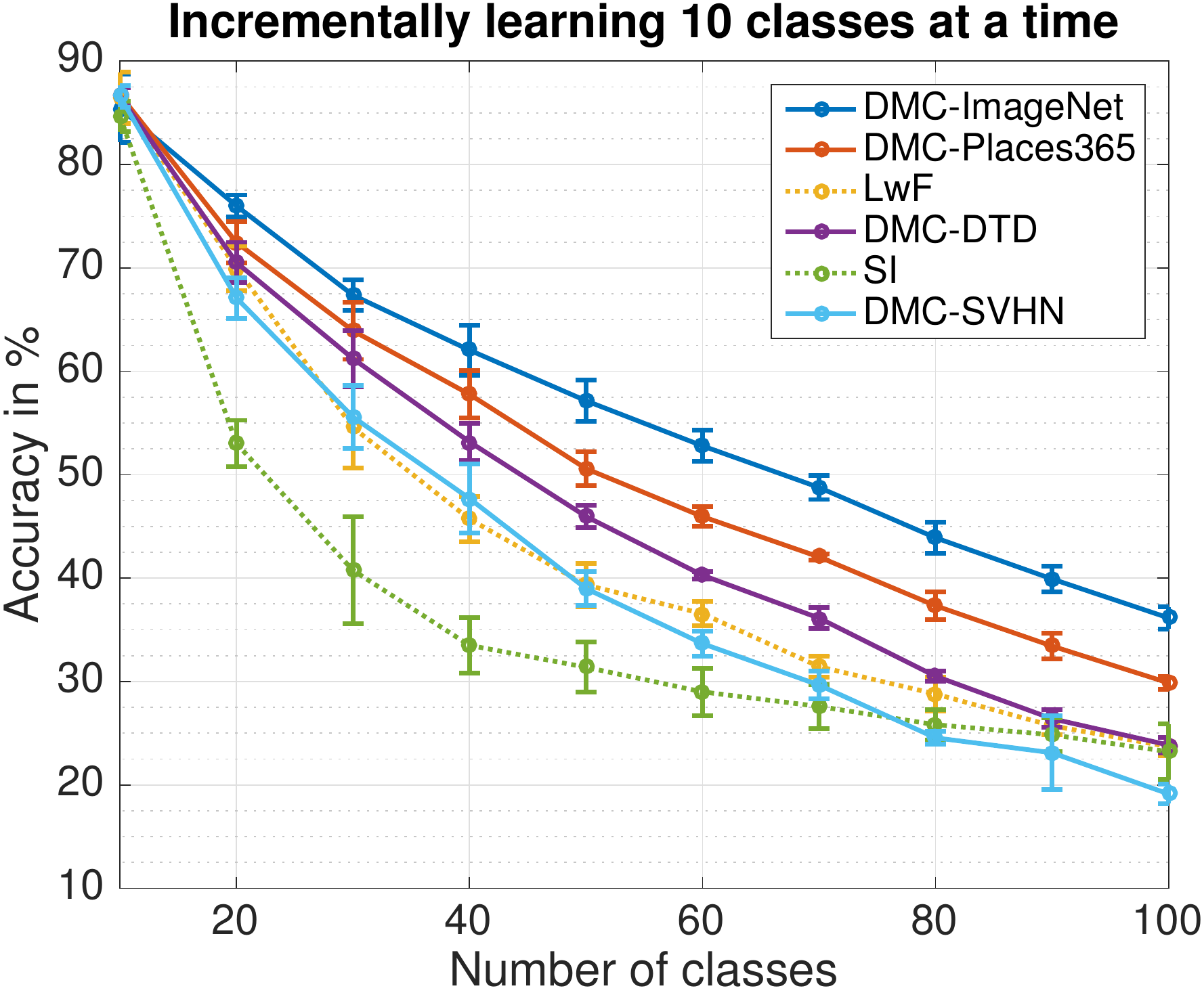}
	\includegraphics[width=0.49\linewidth]{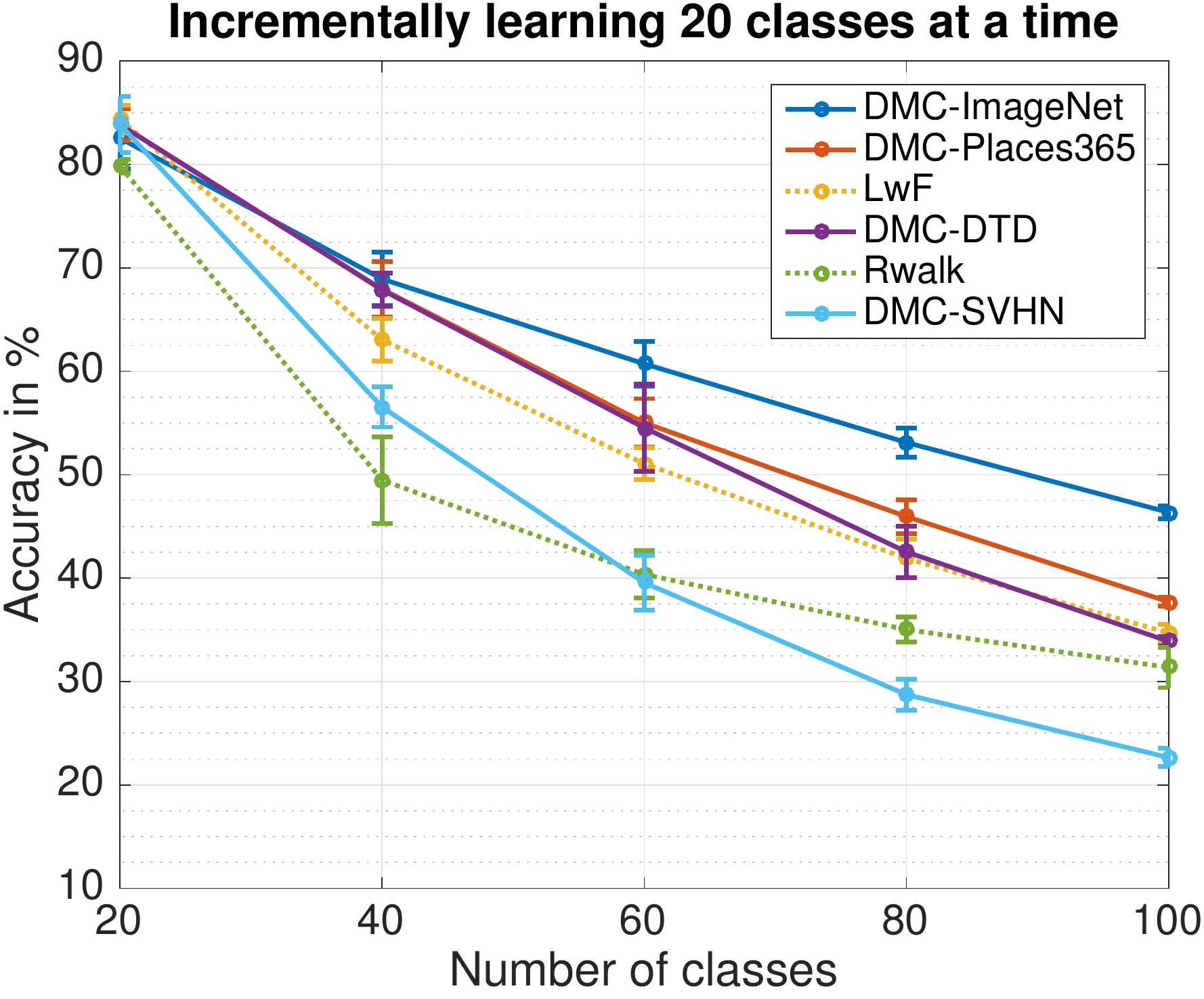}
	\caption{Varying the datasets of auxiliary data used in the consolidation stage on iCIFAR-100 benchmark. Note that using MNIST leads to failure ($\sim$2\% acc.) so we omit the plots.}
	\label{fig:acc-vs-dataset}
\end{figure}

\noindent\textbf{Impact of the distribution of auxiliary data.} Fig. \ref{fig:acc-vs-dataset} shows our empirical study on the impact of the distribution of the auxiliary data by using images from datasets of handwritten digits (MNIST~\cite{lecun1998gradient}), house number digits (SVHN~\cite{netzer2011reading}), texture (DTD~\cite{cimpoi14describing}), and scenes (Places365 \cite{zhou2017places}) as the sources of the auxiliary data. Intuitively, the more diversified and more similar to the target data the auxiliary data is, the better performance we can achieve. Experiments show that usage of overparticular datasets like MNIST and SVHN fails to produce competitive results, but using generic and easily accessible datasets like DTD and Places365 can already outperform the previous state-of-the-art methods. In the applied scenario, one may use the prior knowledge about the target data to obtain the desired auxiliary data from a related domain to boost the performance.

\noindent\textbf{Choices of loss function.} \label{sssec:loss} We compare some common distance metrics used in knowledge distillation in Table \ref{tab:distance} . We observe DMC is generally not sensitive to the loss function chosen, while $L_2$ loss and KD loss~\cite{hinton2015distilling} with $T=2$ performs slightly better than others. As stated in \cite{hinton2015distilling}, both formulations should be equivalent in the limit of a high temperature $T$, so we use $L_2$ loss throughout this paper for its simplicity and stability over various training schedules.

\begin{table}[hbt]
	\vspace{-3pt}
		\centering
		\caption{Average incremental accuracies on CIFAR-100 when $g=20$ and varying distance metrics used in $L_{dd}$.}
		\resizebox{\linewidth}{!}{
			\begin{tabular}{c|c|c|c}
				KD ($T=1$) & KD ($T=2$) & $L_1$ & $L_2$  \\
				\hline
				$46.95 \pm 2.01$ & $58.01 \pm 1.17$ & $57.86 \pm 1.16$ & $58.06 \pm	1.15$
			\end{tabular}
		}
		\label{tab:distance}
			\vspace{-10pt}
	\end{table}

\begin{figure}[htb]
	\begin{center}
		\includegraphics[width=\linewidth]{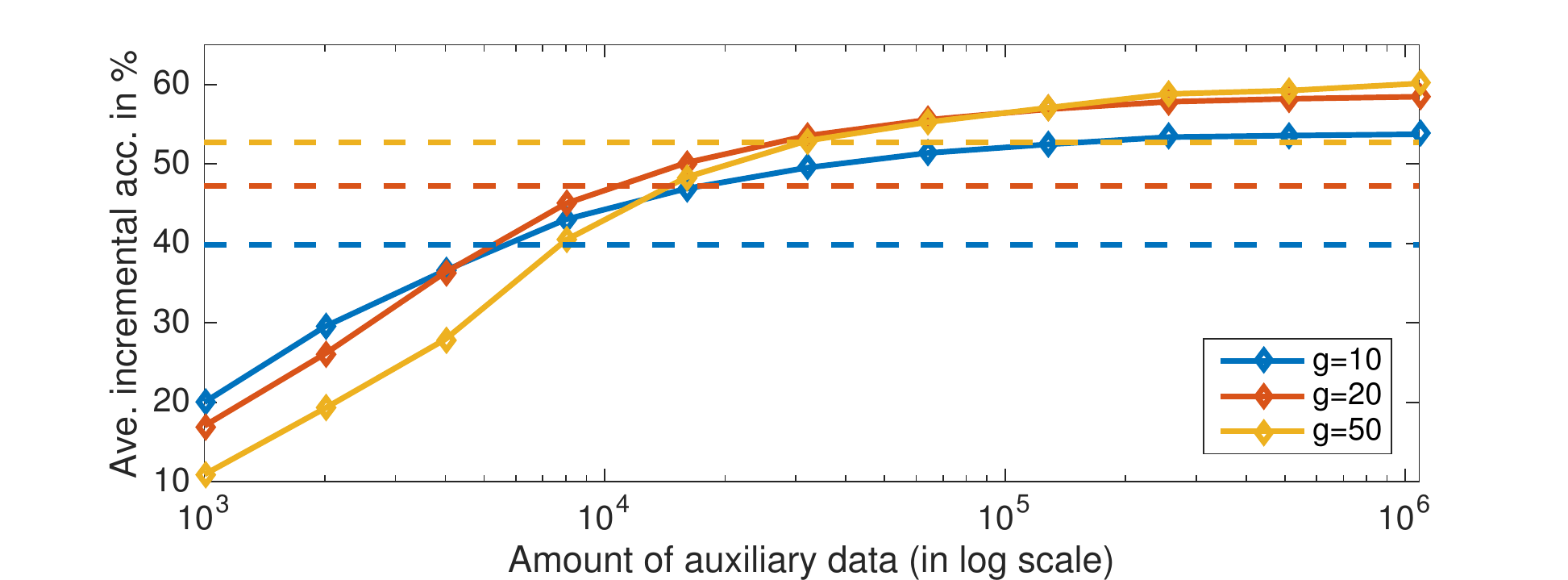}
	\end{center}
			\vspace{-8pt}
	\caption{Average incremental accuracy on iCIFAR-100 with $g=10, 20,50$ classes per group for different the amount of auxiliary data used in the consolidation stage. Dashed horizontal lines represent the performance of the previous state-of-the-art, \ie, \textit{LwF}.}
	\label{fig:cifar100-data}
	\vspace{-8pt}
\end{figure}

\noindent\textbf{Effect of the amount of auxiliary data.} Fig. \ref{fig:cifar100-data} illustrates the effect of the amount of auxiliary data used in consolidation stage. We randomly subsampled $2^k \times 10^3$ images for $k=0,\cdots,9$ from ImageNet32$\times$32 \cite{chrabaszcz2017downsampled}. We report the average of the classification accuracies over all steps of the IL (as in \cite{castro2018end}, the accuracy of the first group is not considered in this average). Overall, our method is robust against the reduction of auxiliary data to a large extent. We can outperform the previous state-of-the-art by just using 8,000, 16,000 and 32,000 unlabeled images ($<3\%$ of full auxiliary data) for $g={10,20,50}$, respectively. Note that it also takes less training time for the consolidated model to converge when we use less auxiliary data.

\noindent\textbf{Experiments with larger images.} We additionally evaluate our method on CUB-200 \cite{WahCUB_200_2011} dataset in IL setting with $g=100$. The network architecture (VGG-16~\cite{simonyan2014very}) and data preprocessing are identical with REWC~\cite{liu2018rotate}. We use BirdSnap \cite{berg2014birdsnap} as the auxiliary data source where we excluded the CUB categories. As shown in Table \ref{tab:cub}, DMC outperforms the previous state-of-the-art~\cite{liu2018rotate} by a considerable margin. This demonstrates that DMC generalizes well to various image resolutions and domains.

\begin{table}[hbt]
	\vspace{-6pt}
	\centering
		\caption{Accuracies on CUB-200 when incrementally learning with $g=100$ classes per group.}
	\small{
		\begin{tabular}{c|c|c|c}
			Methods  & Old Classes  & New Classes & Average Accuracy  \\
			\hline
			EWC~\cite{kirkpatrick2017overcoming}  & 42.3  & 48.6  & 45.3   \\
			\hline
			REWC~\cite{liu2018rotate}  & 53.3  & 45.2  & 48.4   \\
			\hline
			\textbf{Ours}  & $\mathbf{54.70}$ & $\mathbf{57.56}$ & $\mathbf{55.89}$
		\end{tabular}
	}

	\label{tab:cub}
	\vspace{-10pt}
\end{table}

\begin{table*}[htbp]
	\centering
	\caption{VOC 2007 test per-class average precision (\%) when incrementally learning $10+10$ classes.}
\label{tab:pascal_10}%
	\setlength\tabcolsep{3pt}
\resizebox{\linewidth}{!}{
	\begin{tabular}{|c|c|c|c|c|c|c|c|c|c|c|c|c|c|c|c|c|c|c|c|c|c|}
		\hline
	Method & \begin{sideways}aero\end{sideways} & \begin{sideways}bike\end{sideways} & \begin{sideways}bird\end{sideways} & \begin{sideways}boat\end{sideways} & \begin{sideways}bottle\end{sideways} & \begin{sideways}bus\end{sideways} & \begin{sideways}car\end{sideways} & \begin{sideways}cat\end{sideways} & \begin{sideways}chair\end{sideways} & \begin{sideways}cow\end{sideways} & \begin{sideways}table\end{sideways} & \begin{sideways}dog\end{sideways} & \begin{sideways}horse\end{sideways} &  \begin{sideways}mbike\end{sideways} & \begin{sideways}person\end{sideways} & \begin{sideways}plant\end{sideways} & \begin{sideways}sheep\end{sideways} &
	\begin{sideways}sofa\end{sideways} & \begin{sideways}train\end{sideways} &  
	\begin{sideways}tv\end{sideways} & mAP \\
	\hline
	Class 1-10 & 76.8 & 78.1 & 74.3 & 58.9 & 58.7 & 68.6 & 84.5 & 81.1 & 52.3 & 61.4 & - & - & - & - & - & - & - & - & - & - & -\\

Class 11-20 & - & - & - & - & - & - & - & - & - & - & 66.3 & 71.5 & 75.2 & 67.7 & 76.4 & 38.6 & 66.6 & 66.6 & 71.1 & 74.5 & -\\

Oracle & 77.8 & 85.0 & 82.9 & 62.1 & 64.4 & 74.7 & 86.9 & 87.0 & 56.0 & 76.5 & 71.2 & 79.2 & 79.1 & 76.2 & 83.8 & 53.9 & 73.2 & 67.4 & 77.7 & 78.7 & 74.7\\
\hline
Adapted Shmelkov \etal \cite{shmelkov2017incremental} & 67.1 & 64.1 & 45.7 & 40.9 & 52.2 & 66.5 & 83.4 & 75.3 & 46.4 & 59.4 & 64.1 & 74.8 & 77.1 & 67.1 & 63.3 & 32.7 & 61.3 & 56.8 & 73.7 & 67.3 & 62.0 \\

\pname - exclusive aux. data & 68.6 & 71.2 & 73.1 & 48.1 & 56.0 & 64.4 & 81.9 & 77.8 & 49.4 & 67.8 & 61.5 & 67.7 & 67.5 & 52.2 & 74.0 & 37.8 & 63.0 & 55.5 & 65.3 & 72.4 & 63.8 \\

Inference twice & 76.9 & 77.7 & 74.4 & 58.5 & 58.7 & 67.8 & 84.9 & 77.8 & 52.0 & 65.0 & 67.3 & 69.5 & 70.4 & 61.2 & 76.4 & 39.2 & 63.2 & 62.1 & 72.9 & 74.6 & 67.5 \\

\pname & 73.9 & 81.7 & 72.7 & 54.6 & 59.2 & 73.7 & 85.2 & 83.3 & 52.9 & 68.1 & 62.6 & 75.0 & 69.0 & 63.4 & 80.3 & 42.4 & 60.3 & 61.5 & 72.6 & 74.5 & \textbf{68.3} \\
	\hline
\end{tabular}
}%
\end{table*}%

\begin{table*}[htbp]
	\centering
	\caption{VOC 2007 test per-class average precision (\%) when incrementally learning $19+1$ classes.}
\label{tab:pascal_19}%
	\setlength\tabcolsep{3pt}
\resizebox{\linewidth}{!}{
	\begin{tabular}{|c|c|c|c|c|c|c|c|c|c|c|c|c|c|c|c|c|c|c|c|c|c|}
		\hline
		Method & \begin{sideways}aero\end{sideways} & \begin{sideways}bike\end{sideways} & \begin{sideways}bird\end{sideways} & \begin{sideways}boat\end{sideways} & \begin{sideways}bottle\end{sideways} & \begin{sideways}bus\end{sideways} & \begin{sideways}car\end{sideways} & \begin{sideways}cat\end{sideways} & \begin{sideways}chair\end{sideways} & \begin{sideways}cow\end{sideways} & \begin{sideways}table\end{sideways} & \begin{sideways}dog\end{sideways} & \begin{sideways}horse\end{sideways} &  \begin{sideways}mbike\end{sideways} & \begin{sideways}person\end{sideways} & \begin{sideways}plant\end{sideways} & \begin{sideways}sheep\end{sideways} &
	\begin{sideways}sofa\end{sideways} & \begin{sideways}train\end{sideways} &  
	\begin{sideways}tv\end{sideways} & mAP \\
	\hline
Class 1-19 & 70.6 & 79.4 & 76.6 & 55.6 & 61.7 & 78.3 & 85.2 & 80.3 & 50.6 & 76.1 & 62.8 & 78.0 & 78.0 & 74.9 & 77.4 & 44.3 & 69.1 & 70.5 & 75.6 & - & - \\
Class 20 & - & - & - & - & - & - & - & - & - & - & - & - & - & - & - & - & - & - & - & 68.9 & - \\
Oracle & 77.8 & 85.0 & 82.9 & 62.1 & 64.4 & 74.7 & 86.9 & 87.0 & 56.0 & 76.5 & 71.2 & 79.2 & 79.1 & 76.2 & 83.8 & 53.9 & 73.2 & 67.4 & 77.7 & 78.7 & 74.7\\
\hline

Adapted Shmelkov \etal \cite{shmelkov2017incremental} & 61.9 & 78.5 & 62.5 & 39.2 & 60.9 & 53.2 & 79.3 & 84.5 & 52.3 & 52.6 & 62.8 & 71.5 & 51.8 & 61.5 & 76.8 & 43.8 & 43.8 & 69.7 & 52.9 & 44.6 & 60.2\\

\pname - exclusive aux. data & 65.3 & 65.8 & 73.2 & 43.8 & 57.1 & 73.3 & 83.1 & 79.3 & 45.4 & 74.3 & 55.1 & 82.0 & 68.7 & 62.6 & 74.9 & 42.3 & 65.2 & 67.5 & 67.8 & 64.0 & 65.5 \\

Inference twice & 70.6 & 79.1 & 76.6 & 52.8 & 61.5 & 77.6 & 85.1 & 80.3 & 50.6 & 76.0 & 62.7 & 78.0 & 76.5 & 74.7 & 77.0 & 43.7 & 69.1 & 70.3 & 70.0 & 69.5 & 70.1 \\

\pname & 75.4 & 77.4 & 76.4 & 52.6 & 65.5 & 76.7 & 85.9 & 80.5 & 51.2 & 76.1 & 63.1 & 83.3 & 74.6 & 73.7 & 80.1 & 44.6 & 67.5 & 68.1 & 74.4 & 69.0 & \textbf{70.8} \\
	\hline
\end{tabular}
}%
\end{table*}%

\subsection{Incremental learning of object detectors}
\subsubsection{Experimental setup}
\label{sssec:data}
Following \cite{shmelkov2017incremental}, we evaluate \pname for incremental object detection on PASCAL VOC 2007~\cite{pascal-voc-2007} in the IL setting: there are 20 object categories in the dataset, and we incrementally learn $10 + 10$ classes and $19+1$ classes. The evaluation metric is the standard mean average precision (mAP) on the test set. We use training images from Microsoft COCO \cite{lin2014microsoft} dataset as the source of auxiliary data for the model consolidation stage. Out of 80 object categories in the COCO dataset, we use the 98,495 images that contain objects from the 60 non-PASCAL categories.

\label{sssec:arch}
We perform all experiments using RetinaNet \cite{lin2018focal}, but the proposed method is applicable to other one-stage detectors \cite{law2018cornernet,liu2016ssd, redmon2016you, zhang2018single} with minor modifications. In the $10 + 10$ experiment, we use ResNet-50 \cite{he2016deep} as the backbone network for both 10-class models and the final consolidated 20-class model. In $19+1$ experiment, we use ResNet-50 as the backbone network for the 19-class model as well as the final consolidated 20-class model, and ResNet-34 for the 1-class new model. In all experiments, the backbone networks were pretrained on ImageNet dataset \cite{he2016deep}.
 
 \subsubsection{Experimental results and discussions}
 We compare our method with a baseline method and with the state-of-the-art IL method for object detection by Shmelkov \etal \cite{shmelkov2017incremental}. In the baseline method, denoted by \textit{Inference twice}, we directly run inference for each test image using two specialist models separately and then aggregate the predictions by taking the class that has the highest classification probability among all classes, and use the bounding box prediction of the associated model. The method proposed by Shmelkov \etal \cite{shmelkov2017incremental} is compatible only with object detectors that use pre-computed class-agnostic object proposals (\eg, Fast R-CNN \cite{girshick2015fast}), so we adapt their method for RetinaNet by using our novel anchor boxes selection scheme to determine where to apply the distillation, denoted by \textit{Adapted Shmelkov} \etal \cite{shmelkov2017incremental}. 
 
\noindent\textbf{Learning $10+10$ classes.}  The results are given in Table \ref{tab:pascal_10}. Compared to \textit{Inference twice}, our method is more time- and space-efficient since \textit{Inference twice} scales badly with respect to the number of IL sessions, as we need to store all the individual models and run inference using each one at test time. The accuracy gain of our method over the \textit{Inference twice} method might seem surprising, but we believe this can be attributed to the better representations that were inductively learned with the help of the unlabeled auxiliary data, which is exploited also by many semi-supervised learning algorithms. Compared to \textit{Adapted Shmelkov} \etal \cite{shmelkov2017incremental}, our method exhibits remarkable performance improvement in detecting all classes.

\noindent\textbf{Learning $19+1$ classes.} The results are given in Table \ref{tab:pascal_19}. We observe an mAP pattern similar to the $10+10$ experiment. \textit{Adapted Shmelkov} \etal suffers from degraded accuracy on old classes. Moreover, it cannot achieve good AP on the ``tvmonitor" class. Heavily regularized on 19 old classes, the model may have difficulty learning a single new class with insufficient training data. Our \pname achieves state-of-the-art mAP of all the classes learned, with only half of the model complexity and inference time of \textit{Inference twice}. We also performed the addition of one class experiment with each of the VOC categories being the new class. The behavior for each class is very similar to the ``tvmonitor" case described above. The mAP varies from 64.88\% (for new class ``aeroplane") to 71.47\% (for new class ``person") with mean 68.47\% and standard deviation of 1.75\%. Detailed results are in the supplemental material.

\noindent\textbf{Impact of the distribution of auxiliary data.} The auxiliary data selection strategy that was described in Sec. \ref{sssec:data} would potentially include images that contain objects from target categories. To see the effect of data distribution, we also experimented with a more strict data in which we exclude all the MS COCO images that contain any object instance of 20 PASCAL categories, denoted by \emph{\pname - exclusive aux. data} in Table \ref{tab:pascal_10} and \ref{tab:pascal_19}. This setting can be considered as the lower bound of our method regarding the distribution of auxiliary data. We see that even in such a strict setting, our method outperforms the previous state-of-the-art \cite{shmelkov2017incremental}. This study also implies that our method can benefit from auxiliary data from a similar domain.

\noindent\textbf{Consolidating models with different base networks.} As mentioned in Sec. \ref{sssec:arch}, originally we used different base network architectures for the two specialist models in $19+1$ classes experiment. As shown in Table \ref{tab:vary_arch}, we also compare the case when using ResNet-50 backbone for both the 19-class model and the 1-class model. We observed that ResNet-50 backbone does not work as well as ResNet-34 backbone, which could result from overfitting of the deeper model to the training data of the new class and thus it fails to produce meaningful distillation targets in the model consolidation stage. However, since our method is architecture-independent, it offers the flexibility to use any network architecture that fits best to the current training data.

\begin{table}[hbt]
    \centering
    \caption{VOC 2007 test mAP (\%) when using different network architectures for the old and new model, respectively. Classes 1-19 are the old classes, and class 20 (tvmonitor) is the new one.}
    \label{tab:vary_arch}
    \resizebox{\linewidth}{!}{
    \begin{tabular}{c|c|c|c}
      Model  & Old Classes & New Class & All Classes  \\
      \hline
      Class 1-19 (ResNet-50)  & 70.8 & - & - \\
      \hline
      Class 20 (ResNet-34)  & - & 68.9 & - \\
      Consolidated &70.9 &69 &70.8 \\
      \hline
      Class 20 (ResNet-50) & - & 59.0 & -\\
      Consolidated &70.2 &57.9 & 69.9\\
    \end{tabular}
    }
\vspace{-15pt}
\end{table}
\section{Conclusion}
\label{sec:conclusion}
In this paper, we present a novel class-incremental learning paradigm called \pname. With the help of a novel double distillation training objective, \pname does not require storage of any legacy data; it exploits readily available unlabeled auxiliary data to consolidate two independently trained models instead. \pname outperforms existing non-exemplar-based methods for incremental learning on large-scale image classification and object detection benchmarks by a significant margin. \pname is independent of network architectures and thus it is applicable in many tasks. 

Future directions worth exploring include: theoretically characterize how the ``similarity" between the unlabeled auxiliary data and target data affects the IL performance; 2) continue the study on using of exemplars of old data with \pname (presented in supp. material), in terms of exemplar selection scheme and rehearsal strategies; 3) generalize \pname to consolidate multiple models at one time; 4) extend \pname to other applications where consolidation of deep models is beneficial, \eg, taking ensemble of models trained with the same or partially overlapped sets of classes.

\textbf{Acknowledgments.} This work was started during internship at Samsung Research America and later continued at USC. We also acknowledge the support of NVIDIA Corporation with the donation of a Titan X Pascal GPU.

{\small
\bibliographystyle{ieee}
\bibliography{WeeklyReportRef}
}

\newpage
\clearpage
\section*{Appendix Overview}
In this supplemental document, we provide additional detailed experimental results and analyses of the proposed method, Deep Model Consolidation (DMC), for class-incremental learning.

\appendix

\section{Detailed experimental results of \pname for object detection}
In the experiments of \pname for incremental learning of object detectors, we incrementally learn $19+1$ classes using RetinaNet \cite{lin2018focal}. In the main paper, we presented the results of adding ``tvmonitor" class as the new class. Here, we show the results of addition of one class experiment with each of the VOC categories being the new class in Table \ref{tab:pascal_19}, where \textit{Old Model} denotes the 19-class detector trained on the old 19 classes, \textit{New Model} denotes the 1-class detector trained on the new class and \textit{DMC} denotes the final consolidated model that is capable of detecting all the 20 classes.  Per-class average precisions on the entire test set of PASCAL VOC 2007 \cite{pascal-voc-2007} are reported.

\section{Effect of the amount of auxiliary data for object detection} 
We studied the effect of the amount of auxiliary data for DMC for image classification task. To see how the amount of auxiliary data affects the final performance in the incremental learning of object detection, we performed additional experiments on PASCAL VOC 2007 with the $10+10$ classes setting. 
We randomly sampled $1/2$, $1/4$ and $1/8$ of the full auxiliary data from Microsoft COCO dataset~\cite{veit2016coco} for consolidation. As shown in Table \ref{tab:vary_amount_of_aux_data}, with just $1/8$ of full data, \emph{i.e.}, 12.3k images, \pname can still outperform the state-of-the-art, which demonstrates its robustness and efficiency in the detection task as well.

\begin{table}[hbt]
    \centering
    \caption{Varying the amount of auxiliary data in the consolidation stage. VOC 2007 test mAP (\%) are shown, where classes 1-10 are the old classes, and classes 11-20 are the new ones.}
    \label{tab:vary_amount_of_aux_data}
    \resizebox{\linewidth}{!}{
    \begin{tabular}{c|c|c|c}
       Model  & Old Classes & New Classes & All Classes  \\
       \hline
       All auxiliary data  & 70.53 & 66.16 & 68.35 \\
       \hline
       $1/2$ of auxiliary data  & 69.79 & 66.06 & 67.93 \\
       \hline
       $1/4$ of auxiliary data & 70.2 & 64.67 & 67.44 \\
       \hline
       $1/8$ of auxiliary data & 66.77 & 62.71 & 64.74 \\
    \end{tabular}
    }
    \vspace{-2pt}
\end{table}

\section{Implementation and training details}
We implement DMC with PyTorch \cite{paszke2017automatic} library. 

\textbf{Training details for the image classification experiments.} Following iCaRL~\cite{rebuffi2017icarl}, we use a 32-layers ResNet \cite{he2016deep} for all experiments and the model weights are randomly initialized. When training the individual specialist models, we use SGD optimizer with momentum for 200 epochs. In the consolidation stage, we train the network for 50 epochs. The learning rate schedule for all the experiments is same, \emph{i.e.}, it starts with 0.1 and reduced by 0.1 at 7/10 and 9/10 of all epochs. For all experiments, we train the network using mini-batches of size 128 and a weight decay factor of $1\times10^{-4}$ and momentum of 0.9. We apply the simple data augmentation for training: 4 pixels are padded on each side, and a $32\times32$ crop is randomly sampled from the padded image or its horizontal flip.

\textbf{Training details for the object detection experiments.} We resize each image so that the smaller side has 640 pixels, keeping the aspect ratio unchanged. We train each model for 100 epochs and use Adam \cite{kingma2015adam} optimizer with learning rate $1\times10^{-3}$ on two NVIDIA Tesla M40 GPUs simultaneously, with batch size of 12. Random horizontal flipping is used for data augmentation. Standard non-maximum suppression (NMS) with threshold 0.5 is applied for post-processing at test time to remove the duplicate predictions. For each image, we select 64 anchor boxes for DMC training. Empirically we found selecting more anchor boxes (128, 256 etc.) did not provide further performance gain. The $\lambda$ is set to 1.0 for all experiments.

\textbf{Hyperparameters used for the baseline methods.}
We report results of \textit{EWC++}~\cite{chaudhry2018riemannian}, \textit{SI}~\cite{zenke2017continual},  \textit{MAS}~\cite{aljundi2018memory} and \textit{RWalk}~\cite{chaudhry2018riemannian} on iCIFAR-100 benchmark in the main paper. Table \ref{tab:hyperparamter} summarizes the hyperparamter $\hat{\lambda}$ that controls the strength of regularization used in each experiment, and they are picked based on a held-out validation set.

\begin{table}[hbt]
    \centering
    \begin{tabular}{c|c|c|c|c}
         Methods & $g=5$  & $g=10$  & $g=20$ & $g=50$   \\
         \hline
         EWC++~\cite{chaudhry2018riemannian} & 10  &10  & 1 & 0.1 \\
         \hline
         SI~\cite{zenke2017continual} & 0.01  & 0.05  &  0.01 & 0.01 \\
         \hline
         MAS~\cite{aljundi2018memory} & 0.1  & 0.1  & 0.001 & 0.0001  \\
         \hline
         RWalk~\cite{chaudhry2018riemannian} & 5  & 1  & 1 & 0.1 \\
    \end{tabular}
    \caption{$\hat{\lambda}$ used in when incrementally learning $g$ classes at a time on iCIFAR-100 benchmark.}
    \label{tab:hyperparamter}
\end{table}

\section{Preliminary experiments of adding exemplars}
While DMC is realistic in applied scenarios due to its scalablity and immunity to copyright and privacy issues, we additionally tested our method in the scenario where we are allowed to store some exemplars from the old data with a fixed budget when learning the new classes. Suppose we are incrementally learning a group of $g$ classes at a time, With the same total memory budget $K=2000$ as in iCaRL \cite{rebuffi2017icarl}, we fill the exemplar set by randomly sampling $\lfloor \frac{K}{g} \rfloor$ training images from each class when we learn the first group of classes; then every time we learn $g$ more classes with training data $\mathcal{D}_{new} = [X^{gi},\cdots, X^{g(i+1)-1}]$ in the $i$-th incremental learning session, we augment the exemplar set by $\lfloor \frac{K}{gi} \rfloor$ randomly sampled training images of the new classes, and we fine-tune the consolidated model using these exemplars for 15 epochs with a small learning rate of $1 \times 10^{-3}$. After fine-tuning, we reduce the size of the exemplar set by keeping $\lfloor \frac{K}{g(i+1)} \rfloor$ exemplars for each class. We refer to this variant of our method as DMC+. We validate the effectiveness of DMC+ on the iCIFAR-100 benchmark, and Table \ref{tab:exemplar} summarizes the results as the average of the classification accuracies over all steps of the incremental training (as in \cite{castro2018end}, the accuracy of the first group is not considered in this average). We can get comparable performance to iCaRL in all settings. Note that we also tried the herding algorithm to select the exemplars as in iCaRL, but we did not observe any notable improvement.

\begin{figure}[htbp]
	\begin{center}
	\subfigure[DMC+]{\includegraphics[width=0.48\linewidth]{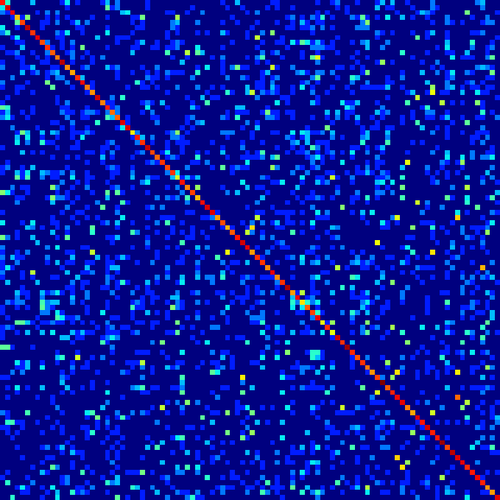}}
	\subfigure[iCaRL]{
	\label{fig:cm_icarl}
	\includegraphics[width=0.48\linewidth]{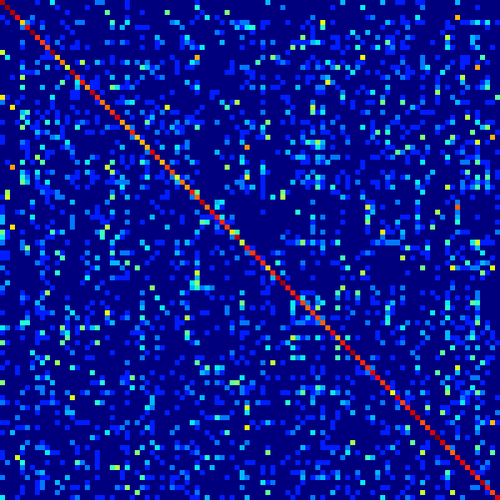}}
	\end{center}
	\caption{Confusion matrices of exemplar-based methods on iCIFAR-100 when incrementally learning 10 classes in a group. The element in the $i$-th row and $j$-th column indicates the percentage of samples with ground-truth label $i$ that are classified into class $j$. Fig. \ref{fig:cm_icarl} is from \cite{rebuffi2017icarl}. (Best viewed in color.)}
	\label{fig:cifar100-cm-exp}
\end{figure}

The confusion matrices comparison between DMC+ and iCaRL \cite{rebuffi2017icarl} is shown in Fig. \ref{fig:cifar100-cm-exp}, and we find: 1) fine-tuning with exemplars can indeed further reduce the intrinsic bias in the training;  2) our DMC+ is on a par with iCaRL, even though we use naive random sampling rather than the more expensive herding \cite{rebuffi2017icarl} approach to select exemplars. 

These preliminary results demonstrate that DMC may also hold promise for exemplar-based incremental learning, and we would like to further study the potential improvement of DMC+, \eg in terms of exemplar selection scheme and rehearsal strategies.

\begin{table}[h!]
    \centering
    \caption{Average incremental accuracies when adding the exemplars of old classes. iCaRL~\cite{rebuffi2017icarl} with the same memory budget is compared. Results of incremental learning with $g= 5,10,20,50$ classes at a time on iCIFAR-100 benchmark are reported.}
    \label{tab:exemplar}
    \resizebox{\linewidth}{!}{
    \begin{tabular}{c|c|c|c|c}
       $g$  & 5  & 10 & 20 & 50 \\
       \hline
       iCaRL  & $\mathbf{57.8 \pm 2.6}$ & $\mathbf{60.5 \pm 1.6}$ & $62.0 \pm 1.2$ & $61.8 \pm 0.4$  \\
       \hline
       DMC+  & $56.78 \pm 0.86$ & $59.1 \pm 1.4$ & $\mathbf{63.2 \pm 1.3}$ & $\mathbf{63.1 \pm 0.54}$
    \end{tabular}
    }
\end{table}

\section{Preliminary experiments of consolidating models with common classes}
 The original DMC assumes the two models to be consolidated are trained with distinct sets of classes, but it can be easily extended to the case where we have two models that are trained with partially overlapped set of classes. We first normalize the logits produced by the two models as Eq. 4 in the main paper. We then set the double distillation regression target as the follows: for the common classes, we take the mean of normalized logits from the two model; for each of the other classes, we take the normalized logit from the corresponding specialist model that was trained with this class.
 
 Below we present a preliminary experiment on CIFAR-100 dataset in this setting, where we have separately trained two 55-class classifiers for \textit{Class 1-55} and \textit{Class 46-100}, respectively, where 10 classes (\textit{Class 46-55}) are in common. The results are shown in Table \ref{tab:common}. For the common classes, DMC can be considered as an ensemble learning method, where at least the accuracy of the weaker model is maintained; for learning the rest of classes, it does not exhibit catastrophic forgetting or intransigence. This shows that DMC is promisingly extensible to the special case of incremental learning with partially overlapped categories.
	
	\begin{table}[h!]
	\vspace{-10pt}
	    \centering
	    \caption{Consolidation of two models with 10 common classes (class 46-55).}
	    \resizebox{\linewidth}{!}{
	    \begin{tabular}{c|c|c|c|c}
	    Model & Class 1-45 & Class 46-55 & Class 56-100 & Class 1-100 \\
	     \hline
	         Model 1 & 73.73 & 80.5 & - & -\\
	         \hline
	         Model 2 & - & 71.6 & 66.47 & - \\
	         \hline
	         Consolidated & 60.76 & 71.7 & 58.09 & 60.65\\
	         \hline
	    \end{tabular}
	    }
	    \label{tab:common}
	    \vspace{-10pt}
	\end{table}

\begin{table*}[htbp]
\centering
	\caption{VOC 2007 test per-class average precision (\%) when incrementally learning $19+1$ classes.}
\label{tab:pascal_19}%
	\setlength\tabcolsep{3pt}
\resizebox{\linewidth}{!}{
	\begin{tabular}{|c|c|c|c|c|c|c|c|c|c|c|c|c|c|c|c|c|c|c|c|c|c|}
		\hline
		Method & \begin{sideways}aero\end{sideways} & \begin{sideways}bike\end{sideways} & \begin{sideways}bird\end{sideways} & \begin{sideways}boat\end{sideways} & \begin{sideways}bottle\end{sideways} & \begin{sideways}bus\end{sideways} & \begin{sideways}car\end{sideways} & \begin{sideways}cat\end{sideways} & \begin{sideways}chair\end{sideways} & \begin{sideways}cow\end{sideways} & \begin{sideways}table\end{sideways} & \begin{sideways}dog\end{sideways} & \begin{sideways}horse\end{sideways} &  \begin{sideways}mbike\end{sideways} & \begin{sideways}person\end{sideways} & \begin{sideways}plant\end{sideways} & \begin{sideways}sheep\end{sideways} &
	\begin{sideways}sofa\end{sideways} & \begin{sideways}train\end{sideways} &  
	\begin{sideways}tv\end{sideways} & mAP \\
\hline
Old Model &
  - &
  78.8 &
  77.4 &
  56.5 &
  60.1 &
  76.4 &
  85.0 &
  80.0 &
  50.0 &
  78.0 &
  69.9 &
  78.3 &
  79.2 &
  74.3 &
  77.3 &
  39.5 &
  66.4 &
  65.7 &
  76.9 &
  74.4 &
  -
   \\
New Model &
  15.8 &
  - &
  - &
  - &
  - &
  - &
  - &
  - &
  - &
  - &
  - &
  - &
  - &
  - &
  - &
  - &
  - &
  - &
  - &
  - &
  -
  \\
DMC &
  16.3 &
  75.9 &
  75.8 &
  52.9 &
  59.5 &
  74.2 &
  84.2 &
  79.1 &
  49.3 &
  73.0 &
  59.9 &
  70.0 &
  75.4 &
  64.8 &
  79.9 &
  40.2 &
  64.1 &
  58.8 &
  69.9 &
  74.3 &
  64.9
   \\
\hline
Old Model &
  69.6 &
  - &
  76.3 &
  60.1 &
  59.8 &
  76.7 &
  85.4 &
  79.6 &
  54.6 &
  75.9 &
  63.7 &
  78.6 &
  79.5 &
  71.5 &
  77.7 &
  44.9 &
  68.0 &
  57.6 &
  77.3 &
  75.5 &
  -
   \\
New Model &
  - &
  70.2 &
  - &
  - &
  - &
  - &
  - &
  - &
  - &
  - &
  - &
  - &
  - &
  - &
  - &
  - &
  - &
  - &
  - &
  - &
  -
  \\
DMC &
  75.8 &
  62.4 &
  75.5 &
  59.6 &
  59.0 &
  76.0 &
  85.6 &
  79.5 &
  53.8 &
  77.0 &
  62.3 &
  78.6 &
  77.6 &
  67.5 &
  80.7 &
  43.5 &
  70.6 &
  57.6 &
  77.3 &
  76.3 &
  69.8
   \\
\hline
Old Model &
  68.9 &
  78.8 &
  - &
  55.9 &
  61.4 &
  70.7 &
  79.9 &
  79.8 &
  50.9 &
  73.6 &
  65.0 &
  77.7 &
  79.3 &
  76.0 &
  77.0 &
  43.1 &
  66.2 &
  66.8 &
  77.3 &
  75.5 &
  -
   \\
New Model &
  - &
  - &
  35 &
  - &
  - &
  - &
  - &
  - &
  - &
  - &
  - &
  - &
  - &
  - &
  - &
  - &
  - &
  - &
  - &
  - &
  -
  \\
DMC &
  69.0 &
  77.9 &
  43.8 &
  54.7 &
  60.1 &
  75.5 &
  84.1 &
  77.5 &
  51.0 &
  71.4 &
  65.4 &
  69.4 &
  69.7 &
  73.5 &
  76.5 &
  40.8 &
  59.9 &
  66.9 &
  77.0 &
  76.2 &
  67.0
   \\
\hline
Old Model &
  76.9 &
  78.3 &
  77.1 &
  - &
  57.9 &
  76.2 &
  85.2 &
  79.8 &
  48.7 &
  76.5 &
  65.6 &
  82.9 &
  76.9 &
  75.1 &
  77.7 &
  40.6 &
  67.7 &
  67.6 &
  76.9 &
  69.5 &
  -
   \\
New Model &
  - &
  - &
  - &
  18.6 &
  - &
  - &
  - &
  - &
  - &
  - &
  - &
  - &
  - &
  - &
  - &
  - &
  - &
  - &
  - &
  - &
  -
  \\
DMC &
  76.6 &
  77.2 &
  75.8 &
  23.4 &
  58.2 &
  77.2 &
  84.4 &
  80.0 &
  48.7 &
  78.5 &
  63.3 &
  82.8 &
  70.3 &
  76.1 &
  80.7 &
  40.8 &
  66.7 &
  64.9 &
  75.5 &
  68.6 &
  68.5
   \\
\hline
Old Model &
  70.5 &
  77.9 &
  77.5 &
  53.5 &
  - &
  76.1 &
  85.6 &
  78.8 &
  51.0 &
  76.2 &
  62.5 &
  77.2 &
  79.1 &
  73.2 &
  77.6 &
  42.5 &
  68.6 &
  68.1 &
  76.6 &
  74.5 &
  -
   \\
New Model &
  - &
  - &
  - &
  - &
  47.7 &
  - &
  - &
  - &
  - &
  - &
  - &
  - &
  - &
  - &
  - &
  - &
  - &
  - &
  - &
  - &
  -
  \\
DMC &
  74.7 &
  76.2 &
  76.4 &
  51.0 &
  37.6 &
  76.9 &
  85.4 &
  79.4 &
  53.0 &
  76.7 &
  64.2 &
  77.9 &
  77.9 &
  73.6 &
  80.4 &
  43.0 &
  68.2 &
  68.4 &
  76.5 &
  75.3 &
  69.6
   \\
\hline
Old Model &
  70.8 &
  77.8 &
  75.2 &
  57.2 &
  60.0 &
  - &
  84.7 &
  79.6 &
  48.3 &
  75.3 &
  68.4 &
  78.8 &
  78.6 &
  75.6 &
  77.3 &
  41.8 &
  69.0 &
  68.0 &
  75.0 &
  73.9 &
  -
   \\
New Model &
  - &
  - &
  - &
  - &
  - &
  46 &
  - &
  - &
  - &
  - &
  - &
  - &
  - &
  - &
  - &
  - &
  - &
  - &
  - &
  - &
  -
  \\
DMC &
  68.7 &
  79.7 &
  73.9 &
  55.6 &
  61.3 &
  53.5 &
  84.9 &
  79.3 &
  49.4 &
  75.8 &
  66.8 &
  78.9 &
  75.6 &
  75.4 &
  80.6 &
  41.6 &
  67.4 &
  66.7 &
  70.0 &
  73.8 &
  68.9
   \\
\hline
Old Model &
  77.5 &
  78.8 &
  74.5 &
  58.1 &
  60.3 &
  74.5 &
  - &
  80.7 &
  49.0 &
  76.0 &
  64.4 &
  77.3 &
  78.7 &
  66.8 &
  77.1 &
  39.0 &
  67.9 &
  67.0 &
  77.1 &
  75.3 &
  -
   \\
New Model &
  - &
  - &
  - &
  - &
  - &
  - &
  76.2 &
  - &
  - &
  - &
  - &
  - &
  - &
  - &
  - &
  - &
  - &
  - &
  - &
  - &
  -
  \\
DMC &
  70.3 &
  76.3 &
  74.0 &
  51.3 &
  60.2 &
  68.2 &
  77.5 &
  80.0 &
  47.1 &
  76.5 &
  61.0 &
  77.4 &
  77.3 &
  59.5 &
  79.9 &
  41.5 &
  66.5 &
  65.0 &
  77.2 &
  74.8 &
  68.1
   \\
\hline
Old Model &
  76.5 &
  79.4 &
  78.1 &
  54.7 &
  60.8 &
  77.2 &
  85.4 &
  - &
  49.6 &
  74.9 &
  65.1 &
  78.5 &
  78.5 &
  74.3 &
  77.8 &
  44.2 &
  67.3 &
  65.1 &
  76.0 &
  74.5 &
  -
   \\
New Model &
  - &
  - &
  - &
  - &
  - &
  - &
  - &
  60.5 &
  - &
  - &
  - &
  - &
  - &
  - &
  - &
  - &
  - &
  - &
  - &
  - &
  -
  \\
DMC &
  75.7 &
  81.0 &
  76.6 &
  51.4 &
  61.9 &
  76.7 &
  84.5 &
  69.8 &
  51.5 &
  74.6 &
  63.6 &
  76.9 &
  69.4 &
  74.6 &
  81.2 &
  43.3 &
  67.0 &
  67.1 &
  77.1 &
  74.2 &
  69.9
   \\
\hline
Old Model &
  78.7 &
  79.6 &
  76.9 &
  57.3 &
  62.2 &
  77.4 &
  80.0 &
  79.5 &
  - &
  75.9 &
  66.6 &
  77.6 &
  79.6 &
  76.5 &
  77.3 &
  43.4 &
  67.3 &
  66.7 &
  77.8 &
  69.3 &
  -
   \\
New Model &
  - &
  - &
  - &
  - &
  - &
  - &
  - &
  - &
  41.9 &
  - &
  - &
  - &
  - &
  - &
  - &
  - &
  - &
  - &
  - &
  - &
  -
  \\
DMC &
  75.8 &
  76.2 &
  76.9 &
  56.7 &
  62.9 &
  76.9 &
  85.4 &
  78.9 &
  38.1 &
  75.1 &
  64.1 &
  78.6 &
  76.0 &
  74.4 &
  80.4 &
  43.0 &
  66.8 &
  62.6 &
  77.8 &
  74.0 &
  70.0
   \\
\hline
Old Model &
  70.8 &
  77.8 &
  76.0 &
  58.1 &
  60.7 &
  78.1 &
  85.0 &
  80.1 &
  47.2 &
  - &
  64.4 &
  77.4 &
  75.3 &
  74.9 &
  80.3 &
  41.7 &
  66.8 &
  64.9 &
  77.1 &
  72.1 &
  -
   \\
New Model &
  - &
  - &
  - &
  - &
  - &
  - &
  - &
  - &
  - &
  30.3 &
  - &
  - &
  - &
  - &
  - &
  - &
  - &
  - &
  - &
  - &
  -
  \\
DMC &
  69.9 &
  75.6 &
  68.1 &
  56.4 &
  60.7 &
  77.2 &
  85.5 &
  79.4 &
  46.4 &
  37.0 &
  65.2 &
  70.0 &
  68.0 &
  74.6 &
  80.4 &
  41.7 &
  59.6 &
  62.8 &
  76.5 &
  72.9 &
  66.4
   \\
\hline
Old Model &
  75.5 &
  80.1 &
  77.1 &
  57.8 &
  61.4 &
  76.6 &
  85.5 &
  80.6 &
  51.1 &
  79.0 &
  - &
  78.6 &
  80.2 &
  75.4 &
  77.1 &
  44.7 &
  68.4 &
  66.7 &
  77.4 &
  74.6 &
  -
   \\
New Model &
  - &
  - &
  - &
  - &
  - &
  - &
  - &
  - &
  - &
  - &
  43.6 &
  - &
  - &
  - &
  - &
  - &
  - &
  - &
  - &
  - &
  -
  \\
DMC &
  75.0 &
  80.8 &
  75.3 &
  54.1 &
  62.6 &
  76.8 &
  85.3 &
  80.6 &
  50.2 &
  77.4 &
  53.9 &
  83.8 &
  77.6 &
  73.4 &
  81.0 &
  45.9 &
  66.8 &
  65.7 &
  75.4 &
  74.8 &
  70.8
   \\
\hline
Old Model &
  76.6 &
  77.8 &
  77.2 &
  57.4 &
  60.6 &
  76.3 &
  84.8 &
  80.9 &
  49.9 &
  77.4 &
  64.5 &
  - &
  77.4 &
  69.3 &
  77.5 &
  43.2 &
  73.7 &
  67.4 &
  76.7 &
  74.7 &
  -
   \\
New Model &
  - &
  - &
  - &
  - &
  - &
  - &
  - &
  - &
  - &
  - &
  - &
  40.3 &
  - &
  - &
  - &
  - &
  - &
  - &
  - &
  - &
  -
  \\
DMC &
  75.2 &
  76.6 &
  74.5 &
  57.1 &
  62.0 &
  74.9 &
  85.4 &
  70.0 &
  51.0 &
  57.8 &
  63.6 &
  52.5 &
  59.2 &
  73.5 &
  79.9 &
  43.1 &
  65.4 &
  66.9 &
  75.1 &
  74.5 &
  66.9
   \\
\hline
Old Model &
  77.3 &
  78.2 &
  77.7 &
  59.4 &
  60.5 &
  77.5 &
  85.2 &
  85.9 &
  48.8 &
  76.6 &
  70.7 &
  76.5 &
  - &
  74.1 &
  77.3 &
  42.1 &
  67.8 &
  68.0 &
  78.7 &
  72.4 &
  -
   \\
New Model &
  - &
  - &
  - &
  - &
  - &
  - &
  - &
  - &
  - &
  - &
  - &
  - &
  52.4 &
  - &
  - &
  - &
  - &
  - &
  - &
  - &
  -
  \\
DMC &
  77.4 &
  76.2 &
  72.1 &
  54.9 &
  63.0 &
  77.5 &
  84.7 &
  79.1 &
  48.0 &
  73.3 &
  68.0 &
  61.5 &
  40.5 &
  71.9 &
  79.5 &
  40.4 &
  65.9 &
  63.0 &
  77.4 &
  73.0 &
  67.4
   \\
\hline
Old Model &
  76.5 &
  77.3 &
  75.7 &
  56.8 &
  60.8 &
  70.5 &
  85.4 &
  79.7 &
  48.6 &
  74.2 &
  62.7 &
  79.3 &
  77.3 &
  - &
  76.9 &
  43.9 &
  68.4 &
  63.3 &
  77.2 &
  76.0 &
  -
   \\
New Model &
  - &
  - &
  - &
  - &
  - &
  - &
  - &
  - &
  - &
  - &
  - &
  - &
  - &
  59.2 &
  - &
  - &
  - &
  - &
  - &
  - &
  -
  \\
DMC &
  68.9 &
  74.2 &
  75.8 &
  55.0 &
  60.1 &
  77.1 &
  84.2 &
  86.0 &
  50.7 &
  75.2 &
  61.3 &
  78.8 &
  70.0 &
  68.2 &
  79.6 &
  46.1 &
  68.1 &
  61.4 &
  75.5 &
  76.1 &
  69.6
   \\
\hline
Old Model &
  77.1 &
  79.7 &
  76.9 &
  59.1 &
  62.3 &
  77.3 &
  85.7 &
  80.2 &
  52.0 &
  77.6 &
  65.0 &
  78.5 &
  80.4 &
  78.2 &
  - &
  44.1 &
  67.5 &
  71.9 &
  78.0 &
  74.1 &
  -
   \\
New Model &
  - &
  - &
  - &
  - &
  - &
  - &
  - &
  - &
  - &
  - &
  - &
  - &
  - &
  - &
  76.4 &
  - &
  - &
  - &
  - &
  - &
  -
  \\
DMC &
  75.8 &
  78.8 &
  75.0 &
  59.7 &
  62.1 &
  77.1 &
  85.5 &
  80.1 &
  51.1 &
  77.0 &
  63.3 &
  77.9 &
  78.1 &
  76.8 &
  78.0 &
  44.7 &
  66.0 &
  69.2 &
  78.1 &
  75.1 &
  71.5
   \\
\hline
Old Model &
  75.5 &
  77.1 &
  75.5 &
  58.9 &
  62.1 &
  77.8 &
  85.8 &
  87.7 &
  44.4 &
  76.6 &
  64.7 &
  78.3 &
  78.7 &
  75.5 &
  77.4 &
  - &
  68.3 &
  67.7 &
  76.6 &
  73.4 &
  -
   \\
New Model &
  - &
  - &
  - &
  - &
  - &
  - &
  - &
  - &
  - &
  - &
  - &
  - &
  - &
  - &
  - &
  35.8 &
  - &
  - &
  - &
  - &
  -
  \\
DMC &
  75.5 &
  74.3 &
  74.4 &
  56.1 &
  61.6 &
  77.9 &
  86.7 &
  87.2 &
  48.0 &
  77.1 &
  64.6 &
  78.0 &
  77.4 &
  73.7 &
  80.3 &
  31.0 &
  67.3 &
  65.8 &
  75.4 &
  73.8 &
  70.3
   \\
\hline
Old Model &
  78.0 &
  77.8 &
  75.3 &
  57.5 &
  61.0 &
  69.6 &
  86.5 &
  79.8 &
  48.5 &
  67.9 &
  62.8 &
  76.8 &
  79.6 &
  74.8 &
  77.5 &
  43.5 &
  - &
  68.0 &
  76.8 &
  74.8 &
  -
   \\
New Model &
  - &
  - &
  - &
  - &
  - &
  - &
  - &
  - &
  - &
  - &
  - &
  - &
  - &
  - &
  - &
  - &
  26 &
  - &
  - &
  - &
  -
  \\
DMC &
  76.3 &
  76.9 &
  73.6 &
  54.3 &
  62.0 &
  73.8 &
  86.1 &
  79.7 &
  48.5 &
  67.0 &
  63.9 &
  75.6 &
  76.3 &
  75.2 &
  80.8 &
  44.4 &
  20.2 &
  68.2 &
  76.4 &
  74.4 &
  67.7
   \\
\hline
Old Model &
  77.6 &
  78.2 &
  76.3 &
  55.0 &
  59.3 &
  70.7 &
  85.8 &
  80.4 &
  50.5 &
  75.4 &
  67.2 &
  83.5 &
  78.7 &
  69.0 &
  77.6 &
  44.4 &
  67.7 &
  - &
  70.1 &
  75.1 &
  -
   \\
New Model &
  - &
  - &
  - &
  - &
  - &
  - &
  - &
  - &
  - &
  - &
  - &
  - &
  - &
  - &
  - &
  - &
  - &
  33.1 &
  - &
  - &
  -
  \\
DMC &
  77.4 &
  82.5 &
  68.5 &
  58.3 &
  61.8 &
  75.2 &
  85.6 &
  78.7 &
  47.1 &
  74.9 &
  63.5 &
  75.6 &
  69.8 &
  73.5 &
  79.4 &
  42.6 &
  65.8 &
  26.1 &
  69.8 &
  74.1 &
  67.5
   \\
\hline
Old Model &
  70.4 &
  79.5 &
  77.1 &
  57.6 &
  60.2 &
  73.6 &
  84.6 &
  80.2 &
  51.0 &
  75.5 &
  65.4 &
  78.7 &
  78.0 &
  75.3 &
  77.7 &
  42.8 &
  69.3 &
  63.6 &
  - &
  73.9 &
  -
   \\
New Model &
  - &
  - &
  - &
  - &
  - &
  - &
  - &
  - &
  - &
  - &
  - &
  - &
  - &
  - &
  - &
  - &
  - &
  - &
  34.9 &
  - &
  -
  \\
DMC &
  73.9 &
  77.4 &
  76.7 &
  56.0 &
  60.8 &
  62.1 &
  83.7 &
  79.7 &
  49.6 &
  76.3 &
  65.7 &
  79.3 &
  73.8 &
  73.2 &
  80.6 &
  40.3 &
  73.3 &
  65.9 &
  37.7 &
  74.5 &
  68.0
   \\
\hline
\end{tabular}%
}
\end{table*}

\section{Enlarged plots}
\begin{figure*}[htbp]
    \centering
	\includegraphics[width=0.49\linewidth]{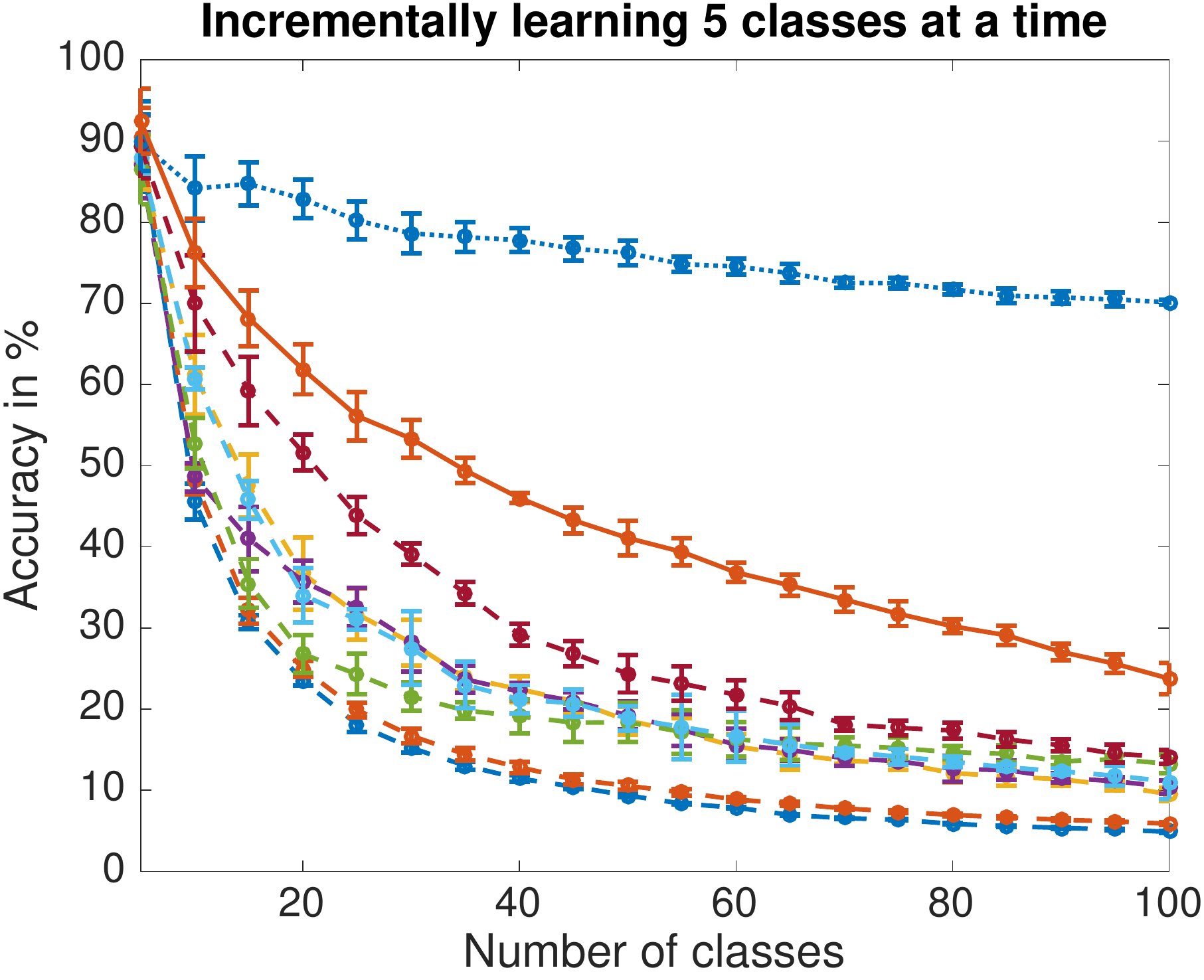}
	\includegraphics[width=0.49\linewidth]{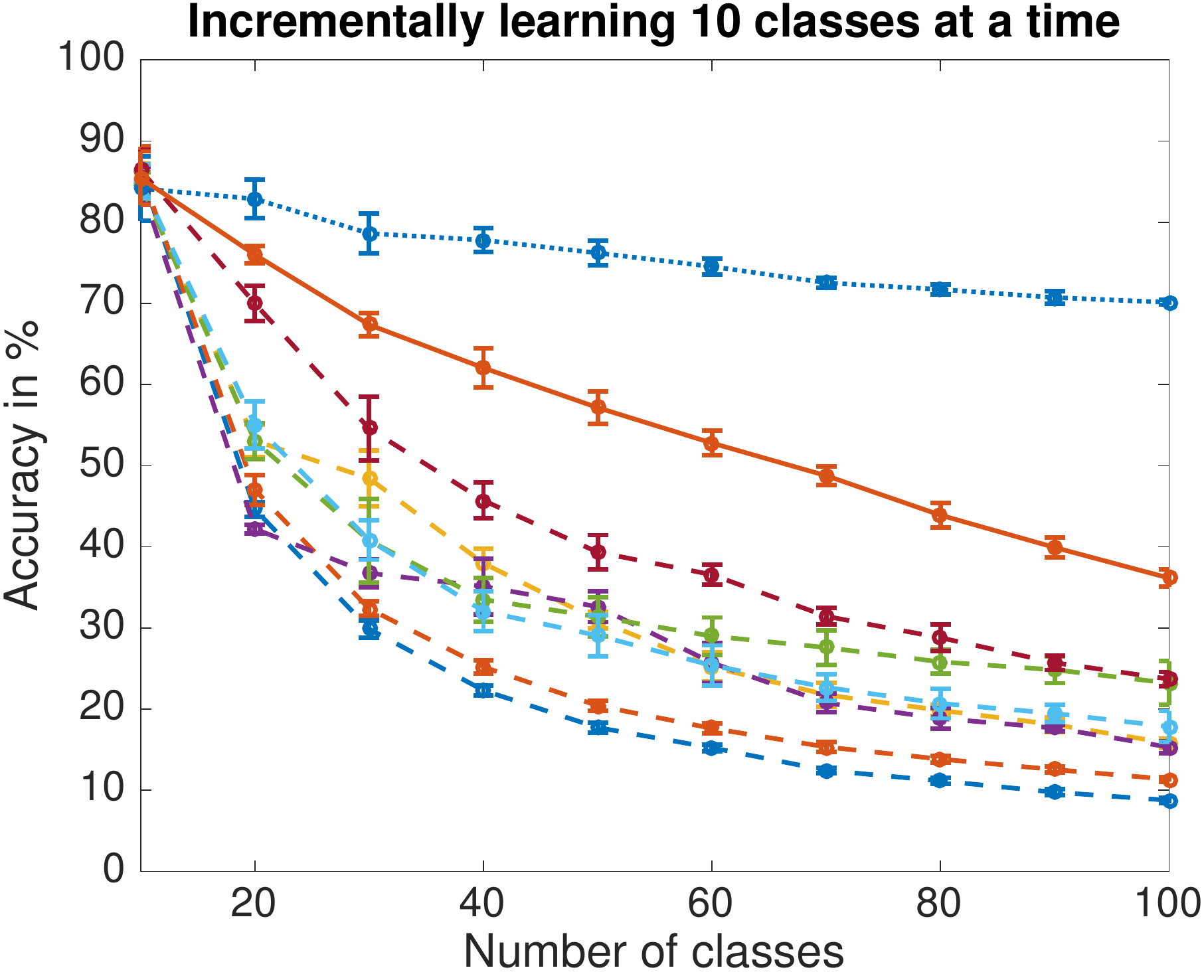}\\
	\vspace{1em}
	\includegraphics[width=0.49\linewidth]{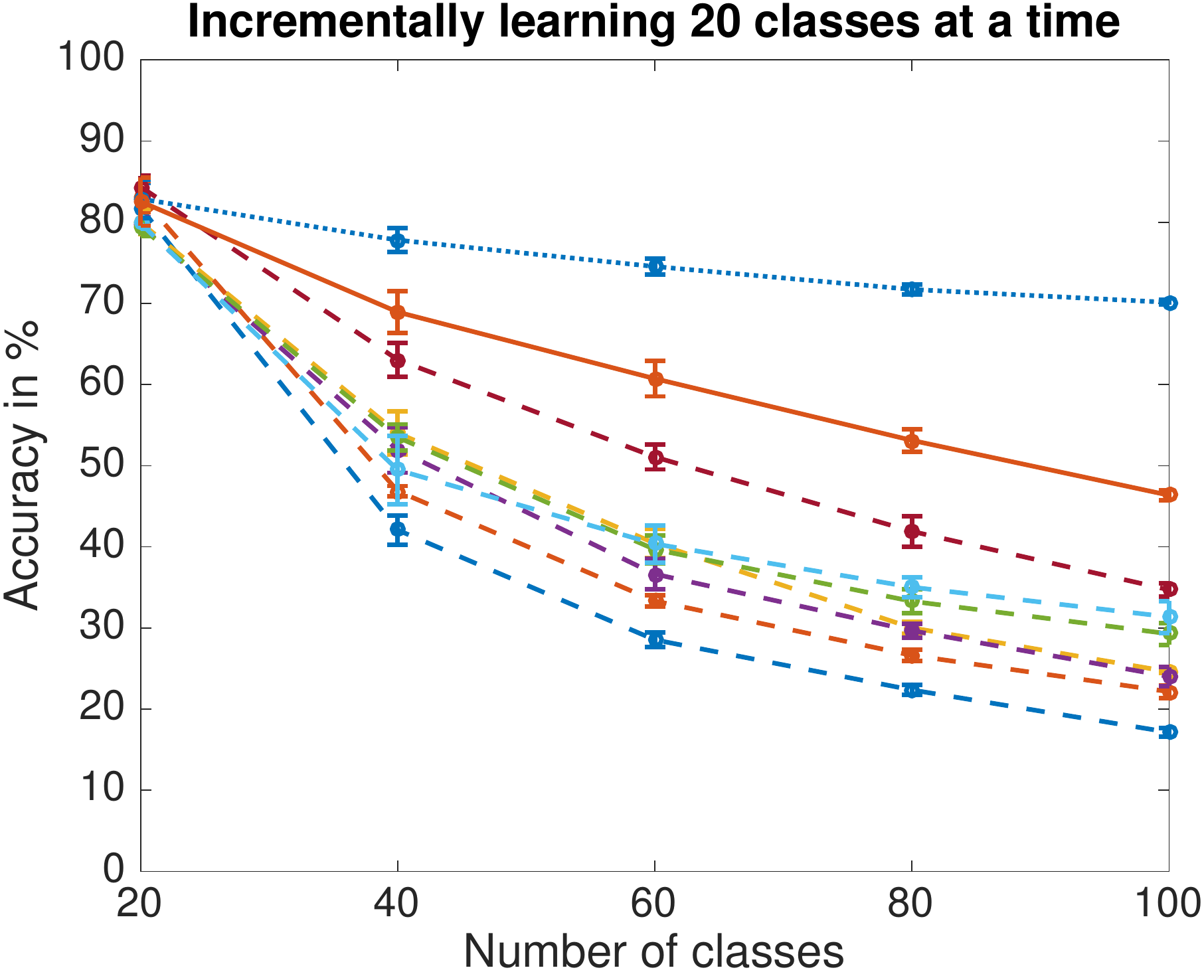}
	\includegraphics[width=0.49\linewidth]{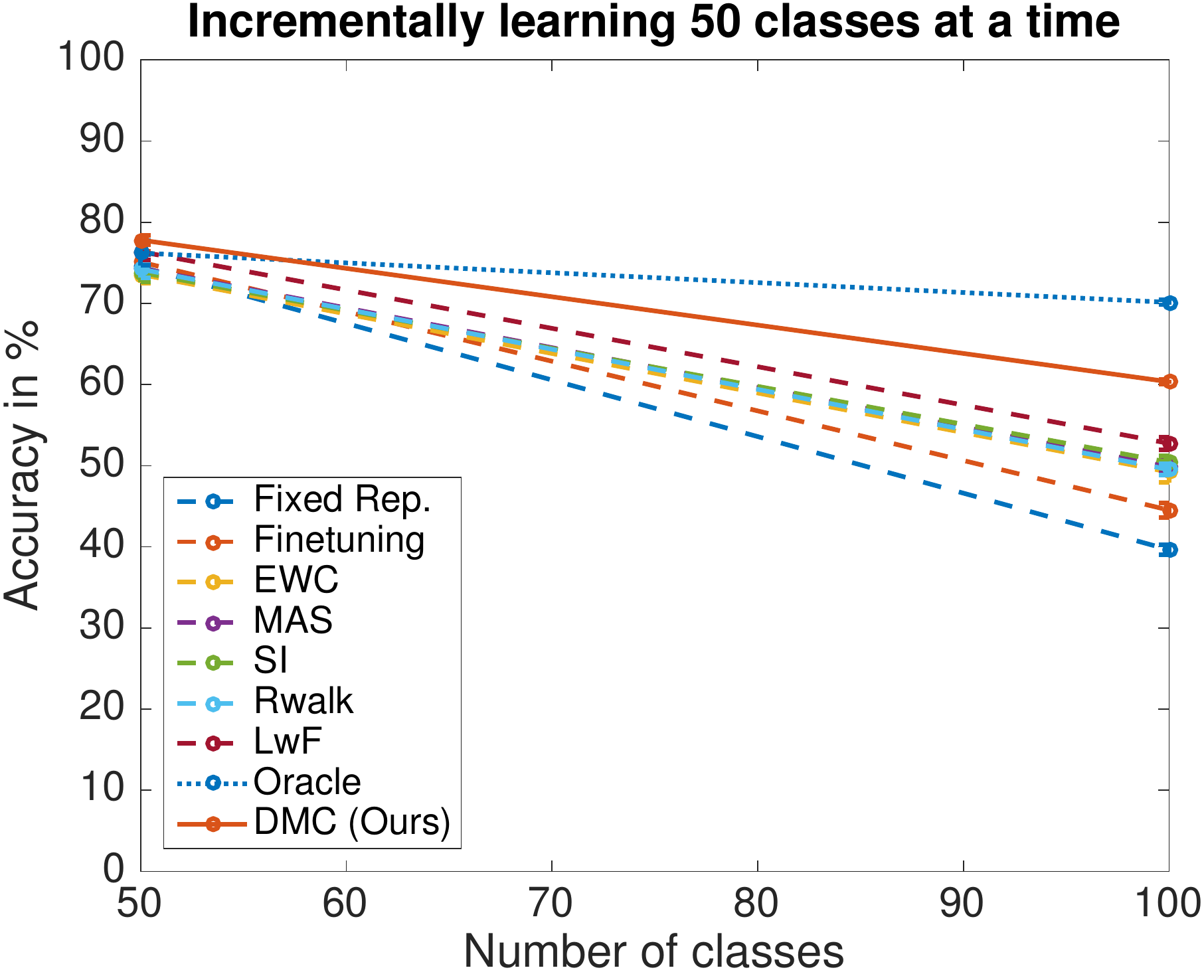}
	\caption{Incremental learning with group of $g=5,10,20,50$ classes at a time on iCIFAR-100 benchmark.}
	\label{fig:large-cifar100}
\end{figure*}

We provide enlarged plots of accuracy curves for iCIFAR-100 ($g=5, 10, 20, 50$) in Fig. \ref{fig:large-cifar100} for better visibility.

\end{document}